%% file: ms.tex
\documentclass[10pt]{article}
\usepackage[margin={2.9cm,2.5cm}, a4paper, headsep=0.25cm]{geometry}

\usepackage{amsmath, amssymb, amsthm, xcolor, supertabular}
\usepackage[hyphens]{url}
\usepackage[colorlinks, citecolor=blue, linkcolor=, urlcolor=]{hyperref}
\usepackage[neverdecrease]{paralist}
\usepackage{mleftright} \mleftright

\usepackage{natbib}
\setcitestyle{round}   
\usepackage{graphicx}
\usepackage[all]{xy}\UseComputerModernTips
\usepackage{placeins}   

\usepackage{microtype}
\DisableLigatures[f]{encoding = *, family = * }

\usepackage{epstopdf}
\usepackage{bm,color,graphicx,mathtools}
\usepackage{caption,subcaption}
\usepackage{dashrule}

\DeclareMathOperator*{\argmin}{arg\,min}

\graphicspath{ {./Figures/}}

\renewcommand*{\equationautorefname}{Equation}
\def\equationautorefname~#1\null{(#1)\null}		

\input{./newcommands.tex}

\newcommand{\arXivOmit}[1]{}

\begin{document}

\title{\bf Ancestral causal learning in high dimensions \\ with a human genome-wide application}

\author{%
    Umberto No\`{e}$^{1}$
	\and
	Bernd Taschler$^{1}$
	\and
	Joachim T\"{a}ger$^{2}$ 
	\and 
    Peter Heutink$^{2}$ 
    \and
	Sach Mukherjee$^{1}$  \vspace{0.1cm} \\
	\small $^{1}$German Center for Neurodegenerative Diseases (DZNE) Bonn, Germany. \\
	\small $^{2}$German Center for Neurodegenerative Diseases (DZNE) T\"{u}bingen, Germany. }

\date{\today}

\maketitle

\begin{abstract}
	\ppara{Abstract:}
	\input{./abstract.tex}
	
	\smallskip
	
	\ppara{Keywords:} 
	causal learning, ancestral causality, human gene expression data, interventional data, supervised learning
	
\end{abstract}

\input{SCL_body.tex}
\FloatBarrier



% ...

\bibliographystyle{abbrvnat}
\addcontentsline{toc}{section}{References}
\bibliography{references}

\newpage
\appendix
\input{supplementary.tex}

\end{document}

%% file: newcommands.tex
\newcommand*{\quark}{\setbox0\hbox{$x$}\hbox to\wd0{\hss$\cdot$\hss}}

\newcommand{\todo}[1]{\bgroup\bfseries\color{red}#1\egroup}

\newcommand*{\arXiv}[1]{URL \bgroup\color{blue}\href{https://arxiv.org/abs/#1}{https://arxiv.org/abs/#1}\egroup}
\newcommand*{\doi}[1]{\bgroup\color{blue}\href{https://dx.doi.org/#1}{doi:#1}\egroup}
\newcommand*{\email}[1]{\bgroup\color{blue}\href{mailto:#1}{#1}\egroup}
\renewcommand*{\url}[1]{\bgroup\color{blue}\href{#1}{#1}\egroup}
\newcommand*{\ppara}[1]{\noindent\textbf{\textsf{#1}}\,\,}


%% file: abstract.tex
We consider learning ancestral causal relationships in high dimensions. Our approach is driven by a supervised learning perspective, with discrete indicators of causal relationships treated as labels to be learned from available data.
We focus on the setting in which some causal (ancestral) relationships are known (via background knowledge or experimental data) and put forward a general approach  that scales to large problems.
This is motivated by problems in human biology which are characterized by high dimensionality and potentially many latent variables. We present a case study involving interventional data from human cells with total dimension $p \! \sim \! 19{,}000$.
Performance is assessed empirically by testing model output against previously unseen interventional data. 
The proposed approach is highly effective and demonstrably scalable to the human genome-wide setting. We consider sensitivity to background knowledge and find that results are robust to nontrivial perturbations of the input information. We consider also the  case, relevant to some applications, where the {\it only} prior information available concerns a small number of known ancestral relationships. 

%% file: SCL_body.tex
\section{Introduction}

We consider the learning of ancestral causal relationships in high dimensions.
Suppose $V = \{ 1 \ldots p \}$ indexes a set of  variables of interest, with $p$ potentially  large.
Our goal is to learn a directed graph $\hat{G}$ comprising vertex set $V$ and edge set $E$, with edges $(i,j) \! \in \! E$ having the interpretation that node $i$ has an ancestral causal influence on node $j$.
Here, an ancestral causal relationship refers to the existence of a directed causal path from the ancestor to the descendant (possibly via latent variables). Such relationships are of particular relevance to experimental studies due to their amenability to empirical verification against experimental  data \citep{Zhang2008CausalGraphs}.
In particular, a change in the variable with index $j$ induced by an intervention on $i$ implies the existence  of such an ancestral path from $i$ to $j$.

Many causal learning methods are based on graphical models, with models based on directed acyclic graphs (DAGs) playing a key  role \citep{Spirtes2001CausationSearch,Pearl2009Causality:Inference}. 
The PC algorithm is a prominent example of such a method \citep{Spirtes2001CausationSearch}. It estimates an equivalence class of DAGs -- encoded as a completed partially directed acyclic graph or CPDAG -- via a series of conditional independence tests. The PC output can in turn be used to estimate bounds on quantitative total causal effects between nodes via an algorithm known as IDA \citep[Intervention calculus when the DAG is Absent;][]{Maathuis2009EstimatingData}. Greedy Interventional Equivalence Search \citep[GIES;][]{Hauser2011CharacterizationGraphs} is a score-based approach that allows for the inclusion of interventional data. 
FCI \citep[Fast Causal Inference;][]{Spirtes2001CausationSearch} and RFCI \citep[Really Fast Causal Inference;][]{Colombo2012LearningVariables} allow for latent variables and learn equivalence classes of ancestral graphs \citep[encoded as partial ancestral graphs or PAGs, see][]{Richardson2002AncestralModels}.
Recently, \cite{Malinsky2017EstimatingSystems} showed how quantitative total causal effects can be obtained from PAGs using the LV-IDA algorithm.

In contrast to these approaches, which are rooted in data-generating models of the causal system, there has been recent work with an emphasis on learning to tell apart causal and non-causal relationships in a data-driven fashion. 
Work in this ``discriminative" direction has included \citet{Lopez-Paz2015TowardsInferenceb}, \citet{Mooij2016DistinguishingBenchmarks} and \citet{Hill2019CausalRegularization}.
Our work follows in this line. Our goals are to put forward a scalable approach for causal learning and investigate whether it can be effective in the challenging context of high-dimensional human genomic data. 

In a nutshell, our approach works as follows. Available information on some ancestral causal relationships (via background knowledge or experimental data) are treated as ``labels" that are combined with a featurization of the data to train a classifier. The fitted classifier is then used to obtain labels (or probabilistic scores) across the entire problem which are intended to encode ancestral causal relationships. 
The output is a directed graph (or associated probabilistic scores) where the presence of an edge $(i,j)$ means that $i$ is inferred to be a causal ancestor of $j$. 
Our work builds on aspects of the  framework presented in \citet{Hill2019CausalRegularization}.
The key methodological difference is that the present paper focuses on high-dimensional settings to which \citet{Hill2019CausalRegularization} does not scale  and on a supervised framework. Furthermore, we present a biological case study using new, genome-wide data. This 
represents a relevant use-case that goes beyond the scale to which their methods can be applied. We include a wide range of experiments investigating behavior in high dimensions and in terms of practically relevant  factors, including sensitivity to incorrect prior information.


Our work is motivated by, and applied to, data from molecular biological experiments on human cells. Such experiments involve measuring molecular quantities  across many thousands of variables (in our example the variables represent gene expression levels and $p \! \sim \! 19{,}000$), including under intervention. The question of interest is to understand causal relationships between these variables. Ancestral relationships are of particular relevance in the context of data of this type  \citep[see e.g.][]{Zhang2008CausalGraphs}. 
Human gene regulation is complex and many of the  underlying processes have been investigated  in considerable mechanistic detail, including via detailed biochemical and biophysical studies. The known set of processes and factors is enormously rich, with many types of variables beyond those included in the available data that are known to influence gene expression levels. These cover, among others, diverse non-coding RNAs (RNAs that are not translated into proteins) as well as proteins and their post-translational modifications  \citep{Mercer2009LongFunctions,Vaquerizas2009AEvolution,Rinn2012GenomeRNAs}. 
From a causal perspective, these mechanisms represent a high dimensional set of latent variables that may influence the observables in complex ways. High dimensionality and the richness of  regulatory influences are characteristic of human  biology and present key challenges for large-scale causal learning in biomedical research. 

Our approach  differs from graphical models-based methods in several key aspects. First, our method copes relatively easily with large numbers of variables, as demonstrated (in terms of both performance and computational efficiency) in our case study. Graphical models-based approaches are theoretically principled and benefit from well-developed causal semantics but the high-dimensional case poses well-known challenges for estimation. That said, it should be noted that while graphical models-based approaches usually aim to model the underlying data-generating process, our approach learns only ancestral edges, which is a more limited form of output. Second, our approach is not hurt by  {\it increasing} dimensionality, since the number of objects being classified grows  with the number of variables and the computations turn out to be very tractable, as we discuss below.
We show empirically using the high-dimensional human data that the performance of our approach is stable as dimension increases up to $p \! \sim \! 19{,}000$. 
Third, our approach allows for cycles in the graph. This stands in contrast to PAG-based ancestral approaches like FCI and RFCI. In biological settings, cycles are common, for example due to feedback regulation in the underlying systems \citep{Alon2006AnCircuits}. Cyclic causation has been discussed in, for example, \citet{Spirtes1995DirectedModels,Richardson1996AGraphs,Hyttinen2001LearningVariables}.


The remainder of the paper is organized as follows. We first introduce notation and describe the methods. We then present empirical results and conclude with a discussion.

\section{Methods}

\subsection{Notation}

\noindent 
{\it Variables and graphs}.
The set of observed variables of interest is indexed by $V \! = \! \{ 1 \ldots p \}$ and the variables themselves are denoted $(X_1, \ldots ,X_p)$.
The variables will often be identified with vertices in a directed graph. Where useful, the vertex and edge sets of a directed graph $G$ are denoted $V(G)$ and $E(G)$, respectively. Occasionally we  overload $G$ to refer also to the corresponding binary adjacency matrix, as will be clear from context. Then, for a graph $G$,  $G_{ij}$ refers to the entry $(i,j)$ of the corresponding adjacency matrix.

\smallskip

\noindent 
{\it Linear indexing}.
To facilitate mapping to the machine learning problem, we use linear indexing of variable pairs. Specifically, an ordered pair $(i,j) \! \in \! V \! \times \! V$ has an associated linear index  $ k \in \mathcal{K} = \{1 \ldots K \}$, where $K$ is the total number of variable pairs of interest. 
Where useful to make the mapping explicit, the linear index corresponding to a pair $(i,j)$ is denoted as $k(i,j)$ and the variable pair corresponding to a linear index $k$ as $(i(k), j(k))$.
The linear indices of variable pairs whose ancestral causal relationships are unknown and of interest are denoted $\mathcal{Q} \subset \mathcal{K}$. That is, for all pairs $\{ (i,j) : k(i,j) \in \mathcal{Q} \}$ we want to learn whether or not $X_i$ has an ancestral causal influence on $X_j$. 

\smallskip

\noindent 
{\it Prior or background information}. 
Some ancestral causal relationships $\mathcal{T}(\Pi) \subset \mathcal{K}$ are known in advance via background knowledge $\Pi$.
In all empirical experiments the sets $\mathcal{T}(\Pi)$ and $\mathcal{Q}$ are disjoint, i.e.\ no prior causal information is available on the pairs $\mathcal{Q}$ of interest.  


\subsection{Ancestral causal learning via a supervised formulation}

Our approach is based on framing the task of learning ancestral causal relationships as a supervised learning problem. We first formulate  the problem in a generic manner without specifying implementation details. This makes it clear how specific combinations of featurization and supervised learning could be used to perform ancestral causal learning. We then describe the particular formulations used in experiments below. 
Our goal is to learn ancestral causal relationships between the  variables $V$.
These relationships are amenable to experimental verification \citep{Zhang2008CausalGraphs, Kocaoglu2017ExperimentalVariables} and are of particular interest for follow-up experiments.

\subsubsection{Problem setting}

The inputs for the given problem setting are: (i) an $n \times p$ data matrix $X$ containing data on $p$ variables indexed $V \! = \! \{ 1 \ldots p \}$; (ii) knowledge $\Pi$ concerning ancestral causal relationships on a subset $\mathcal{T}(\Pi) \! \subset \! V {\times} V$ of variable pairs; and (iii) a set $\mathcal{Q} \! \subset \! V {\times} V$ of variable pairs whose ancestral causal relationships are unknown and of interest. It is assumed that $\mathcal{T}$ and $\mathcal{Q}$ are disjoint.
This corresponds to the assumption that no prior information is available on the causal relationships  of interest, i.e.\ that the causal relationships in $\mathcal{Q}$ are entirely unseen.
Similarly, while no specific assumption is made on the data $X$ (in particular the data need not be i.i.d.), it is assumed that it does not contain interventional data corresponding to the pairs in $\mathcal{Q}$. 

\subsubsection{Featurization}

In order to formulate the problem as a machine learning task, we require a featurization that can be applied to both the pairs $\mathcal{T}$ whose causal status is known from background information $\Pi$ {\it and} the unknown pairs $\mathcal{Q}$.
Let $\phi_k = \phi_k(X) \in \mathbb{R}^d$ generically denote such a featurization of the $p$-dimensional data (for generality this is written as a function of the entire data matrix).
The subscript $k$ indicates that the features are specific to a variable pair $k \in \mathcal{K}$; this is important to allow linking to labels encoding causal status (see below). We consider a specific, very simple featurization below via vectorized bivariate histograms following \citet{Hill2019CausalRegularization}.
Collecting the features over all variable pairs $\mathcal{K}$ gives rise to a $K \times d$ feature matrix $\Phi = [ \phi_1(X) \ldots \phi_K(X)]^{\mathrm{T}}$.

\subsubsection{Learning}

\noindent 
{\it Labels}.
The background information $\Pi$ provides information on ancestral causal relationships for the variable pairs $\mathcal{T}(\Pi)$.
In the specific case of available interventional data, this means that prior interventional experiments reveal, for each pair $k \in \mathcal{T}$, whether or not $i(k)$ is a causal ancestor of $j(k)$ (recall that $i(k), j(k)$ is the ordered  pair corresponding to the linear index $k$). This information is used to form (training) labels $( y_k )_{k \in \mathcal{T}}$. Specifically, 
\begin{equation}
    y_k = 
        \begin{cases}
        1 & \text{if } \text{$i(k)$ is a causal ancestor of $j(k)$ according to $\Pi$}\\
        0 & \text{otherwise.}
        \end{cases}
    \label{eq:labels}
\end{equation}

Collecting together these labels gives a label vector $Y(\Pi) = (y_k)_{k \in \mathcal{T}}$, where the notation emphasizes  that the labels are derived from $\Pi$.

\medskip
\noindent 
{\it The prediction function}.
Thus, labels are available for all vertex pairs $k \! \in \! \mathcal{T}$. For one such pair, the corresponding features are $\phi_k \! \in \!  \mathbb{R}^d$ (the $k^{\mathrm{th}}$ row of the feature matrix $\Phi$). That is, $(\phi_k, y_k)_{k \in \mathcal{T}}$ form (feature, label) couples in the usual supervised learning sense. 
Let $f : \mathbb{R}^d \rightarrow \{ 0,1\}$ denote a prediction function that maps features to a binary label (or probabilistic score, in which case the range of $f$ is $[0,1]$). Assume $f$ can be fully specified by an unknown parameter $\theta$, giving the classification function as 
$f(\, \cdot \, ; \theta)$ (here we use $\theta$ in a general sense to encode all information needed to specify $f$, potentially including parameters, hyper-parameters, model architecture, etc.). 
Estimation of $\theta$ -- and thereby of the prediction function -- is done in a supervised manner via
\begin{equation}
    \hat{\theta}(X,\Pi) = \argmin_{\theta \in \Theta} \sum_{k \in \mathcal{T}(\Pi)} \mathrm{RL}(f(\phi_k(X); \theta), y_k(\Pi)),
    \label{eq:learning}
\end{equation}
\noindent
where $\mathrm{RL}$ denotes generically a loss function with regularization (specific choices correspond to various supervised learning approaches) and $\Theta$ denotes the parameter space. Note that the learning step (\ref{eq:learning}) requires only the variable pairs $\mathcal{T}$ for which the labels are available. 

Now, using $f$,  the estimated label for a new pair $k'$ is 
\begin{equation}
\hat{y}_{k'} = f(\phi_{k'}; \hat{\theta}(X,\Pi))
\label{eq:classifier}
\end{equation}

\noindent
with the notation emphasizing that estimation (of the parameters $\theta$ that in turn specify the function  $f$) is based on the data and background information.

\medskip
\noindent 
{\it The estimate $\hat{G}$}.
Finally, the fitted classifier $f(\, \cdot \, ; \hat{\theta}(X,\Pi))$ is used to obtain labels for all unknown pairs $\mathcal{Q}$ via (\ref{eq:classifier}) using the features $\{ \phi_k \}_{k \in \mathcal{Q}}$. 
This gives an estimate $\hat{G}(X,\Pi)$ with edges specified by
\begin{equation}
    \hat{G}_{ij} = 
    \begin{cases}
        f(\phi_{k(i,j)}; \, \hat{\theta}(X,\Pi)) & \text{if } k(i,j) \notin \mathcal{T}(\Pi)\\
        y_{k(i,j)}(\Pi) & \text{otherwise}
    \end{cases}
    \label{eq:graph_supervised}
\end{equation}

\noindent
where $(i,j)$ are ordered variable pairs.
Note that the overall estimate depends only on the data $X$ and background information $\Pi$, via the parameters of $f$, as estimated via (\ref{eq:learning}). No change is made for pairs $\mathcal{T}$ whose status was known at the outset.


	    

\subsubsection{Correcting prior information}
\label{sec:correction}

In some settings, available information $\Pi$ may not be entirely trustworthy. In this case, it may be desirable to allow the learner to attempt to correct the input labels. We consider doing so by simply applying the fitted classifier to {\it all} points $\mathcal{K}$ (i.e. not only those in $\mathcal{Q}$). This gives an alternative to (\ref{eq:graph_supervised}) with edges estimated as
\begin{equation}
    \hat{G}_{ij}(X, \Pi) = f(\phi_{k(i,j)}; \hat{\theta}(X,\Pi)), \, \forall k \in \mathcal{K}.
\end{equation}

Note that in contrast to (\ref{eq:graph_supervised}) this means that for a pair $k \in \mathcal{T}$ whose causal status is (thought to be) known from background knowledge it may be that the output $\hat{G}_{ij} \neq y_{k(i,j)}$ (``error correction").

\subsection{Computational scaling}
Several computational aspects make the proposed approach highly efficient for large-scale applications. In particular: 

\begin{itemize}
\item The learning step (\ref{eq:learning}) is restricted to the set $\mathcal{T}$ 
which in most applications will be much smaller than the complete set of pairs $\mathcal{K}$. 
If very large amounts of interventional data were available for the training step, (\ref{eq:learning}) could be handled using standard techniques for supervised learning for large data, such as batch-wise, stochastic gradient type methods (these are already used in the neural network examples below). 
\item For problems with large $p$, the estimation of the graph structure (\ref{eq:graph_supervised}), which spans to the complete set of pairs $\mathcal{K}$, involves applying a fitted classifier to a potentially  large number $| \mathcal{Q} | $ of points. This can typically be done very efficiently computationally. 
\item For featurizations that involve dealing with variable pairs, construction of the feature matrix $\Phi$ is $O(p^2)$, however this can usually be parallelized (this is the case for the featurization considered below). 
\end{itemize}

Taken together, the foregoing points mean that the proposed  approach can be effectively  scaled to very high dimensional problems. To give an idea of practical compute times, the largest experiments in the case study below, involving $p\sim 19{,}000$ variables, took roughly 10 minutes to solve on a standard workstation (24 CPU cores and 256GB of RAM). 
Figure~\ref{fig:vary-p-time} in the Appendix shows run-times for our method and various standard algorithms as a function of $p$: the favourable scaling due to the reasons mentioned above can be clearly seen.

\subsection{Specific model formulations used in the experiments}

The overall approach we propose is modular in the sense that, in principle, any featurization $\phi$ could be combined with any predictor $f$ to provide a learner $\hat{G}$. In this Section, we provide some specific formulations that we use in experiments below. 
For featurization, we use a simple bivariate histogram approach following \citet{Hill2019CausalRegularization}. 
In brief, this involves constructing a bivariate histogram for each pair $k$, with $\phi_k$ being obtained as a PCA reduction of vectorized histogram bin counts down to $d=100$ dimensions. 

We consider the following specific classifiers, which we refer to as Scalable Causal Learning (SCL):
\begin{itemize}
    \item {\bf SCL-L1}. Featurization as above, with prediction done using a standard $\ell_1$-regularized generalized linear model, as implemented in \texttt{glmnet} \citep{Friedman2010RegularizationDescent}.
    \item {\bf SCL-NN}. As in SCL-L1, but with prediction performed using  a feedforward neural network (implemented via \texttt{tensorflow} \citep{Abadi2016Tensorflow:Learning}, containing a total of 5 dense layers and 132{,}865 parameters).
\end{itemize}

\section{Results}

\subsection{Overall set-up}

\noindent
{\it Data.} We consider a case study involving  human gene expression data. 
After standard pre-processing and trimming, the data consisted of sequencing-based measurements of gene expression levels of $p=24{,}052$ genes 
in human cells (the neuroblastoma cell line BE(2)-M17)
under intervention by short hairpin RNAs.
The main interventional data consisted of 70 interventions, under each of which all $p$ gene expression levels were measured, and an additional panel  (comprising 35 interventions) that was used solely to define gene-specific thresholds (see below).
Specific genes affected by too many interventions could make the learning problem too easy, we therefore excluded genes affected by half or more of the interventions, leading to a total of approximately $19{,}000$ genes.

We sampled the data to obtain specific problem instances in the following way. A subset of 35 interventions was used to train and test the models (as described above, in terms of the set notation this means $|\mathcal{K}| = 35 \! \times \! p$). The remaining 35 interventional data were used to populate the data matrix $X$ along with the observational data. The intention was to mimic the realistic setting in which the available data contains variation of unspecified source. Note that the interventions included in $X$ are neither used to provide background information $\Pi$ nor to test the models.

The general set-up is that we use some interventional data to train the learners and then test the output against unseen interventional data. A gene $i$ was said to have an ancestral causal effect on gene $j$ if $j$'s expression level under intervention on $i$ was larger or smaller than any measurement (for the same gene $j$) in the panel of additional interventional data. This emphasizes salient changes whilst ensuring an appropriate scale for each gene $j$.
The data were strictly split in the sense that (i) no data used to define the true ancestral relationships 
against which the model output was tested appear in the data matrices $X$ and (ii) in all experiments the pairs $\mathcal{Q}$ on which the model output is tested are disjoint from those pairs $\mathcal{T}$ whose ancestral relationships are provided as auxiliary background information.

\smallskip
\noindent
{\it Comparisons.} 
We compare the proposed method against a panel of existing causal learning methods, including PC, IDA, RFCI and GIES (implemented in the R package \texttt{pcalg} \citep{Kalisch2012CausalPcalg}). 
We obtained total causal effects from the output of RFCI using the LV-IDA method described in \cite{Malinsky2017EstimatingSystems}.
These methods differ in the nature of their output and, as noted at the outset, should not be regarded as direct competitors to our approach. Rather, we include them as important examples of causal learning methods whose output can reasonably be tested with respect to interventional experiments of the kind we consider here. While IDA and RFCI/LV-IDA provide output that can reasonably be directly compared with the interventional data (since they estimate total causal effects), this is not the case for PC and GIES. In the case of PC and GIES we therefore considered an additional transitive closure step to extract ancestral statements from the CPDAG or essential graph output. 

Additionally, we considered some non-causal statistical  methods  including Gaussian graphical models (GGMs; estimated using a shrinkage estimator as implemented in the R package \texttt{corpcor} \citep{Schafer2005AGenomics}), and Pearson and Kendall correlation coefficients. Although non-causal, these approaches are included to provide a set of simple data-driven baselines. 
We note that non-causal methods of this kind remain widely used in many applied settings with scientific goals that are arguably causal (including in bioinformatics), hence it is interesting to consider their empirical behavior with respect to human interventional data. 

In the experiments below, we consider in turn varying a number of key factors, including total problem dimension and the amount of available background information. We go on to assess sensitivity to incorrect background information, where we intentionally induce errors in the input information and examine whether such incorrect background information can be corrected using the ``error correction" approach described above. Finally, we look at a set-up, relevant in several applications, where background information is very sparse in the sense of comprising information on only a small number of known ancestral causal relationships (rather than the entire set of changes under a given intervention). 


\subsection{Varying dimension \texorpdfstring{$p$}{p}}

\begin{figure}
    \centering
    \subcaptionbox{}{
        \hspace{1.2cm}\includegraphics[width=0.7\textwidth]{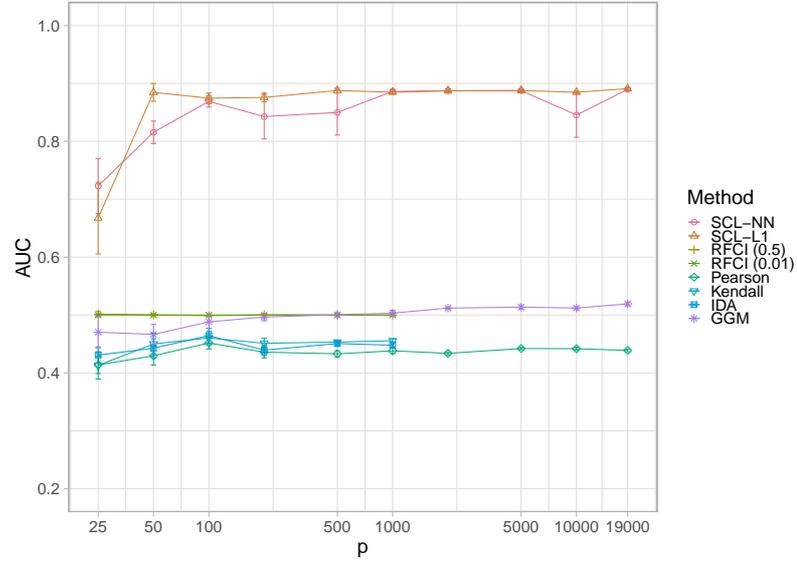}
    }
    \\
    \subcaptionbox{$p = 100$}{
        \includegraphics[width=0.3\textwidth]{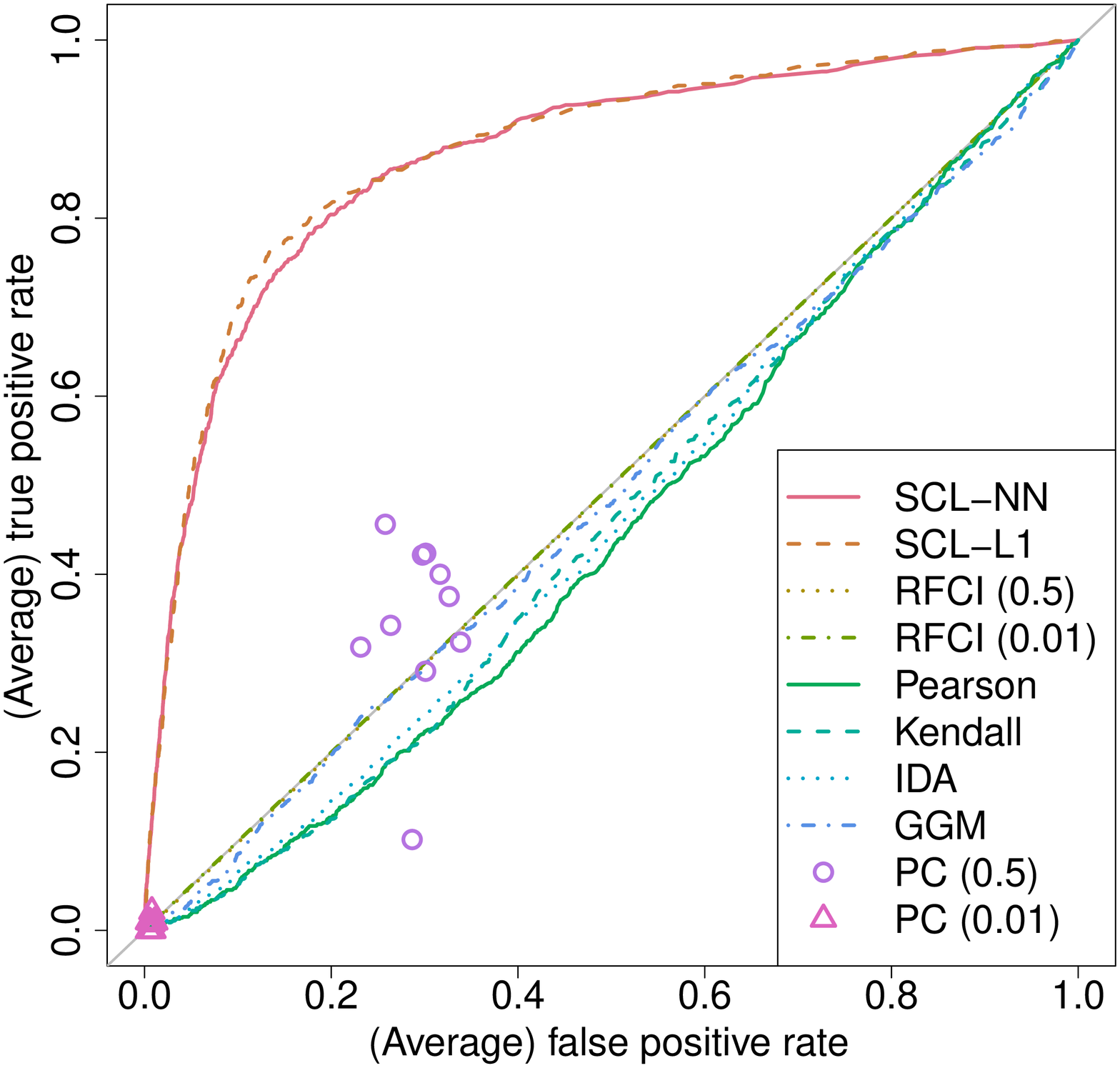}
    }
    \subcaptionbox{$p = 1000$}{
        \includegraphics[width=0.3\textwidth]{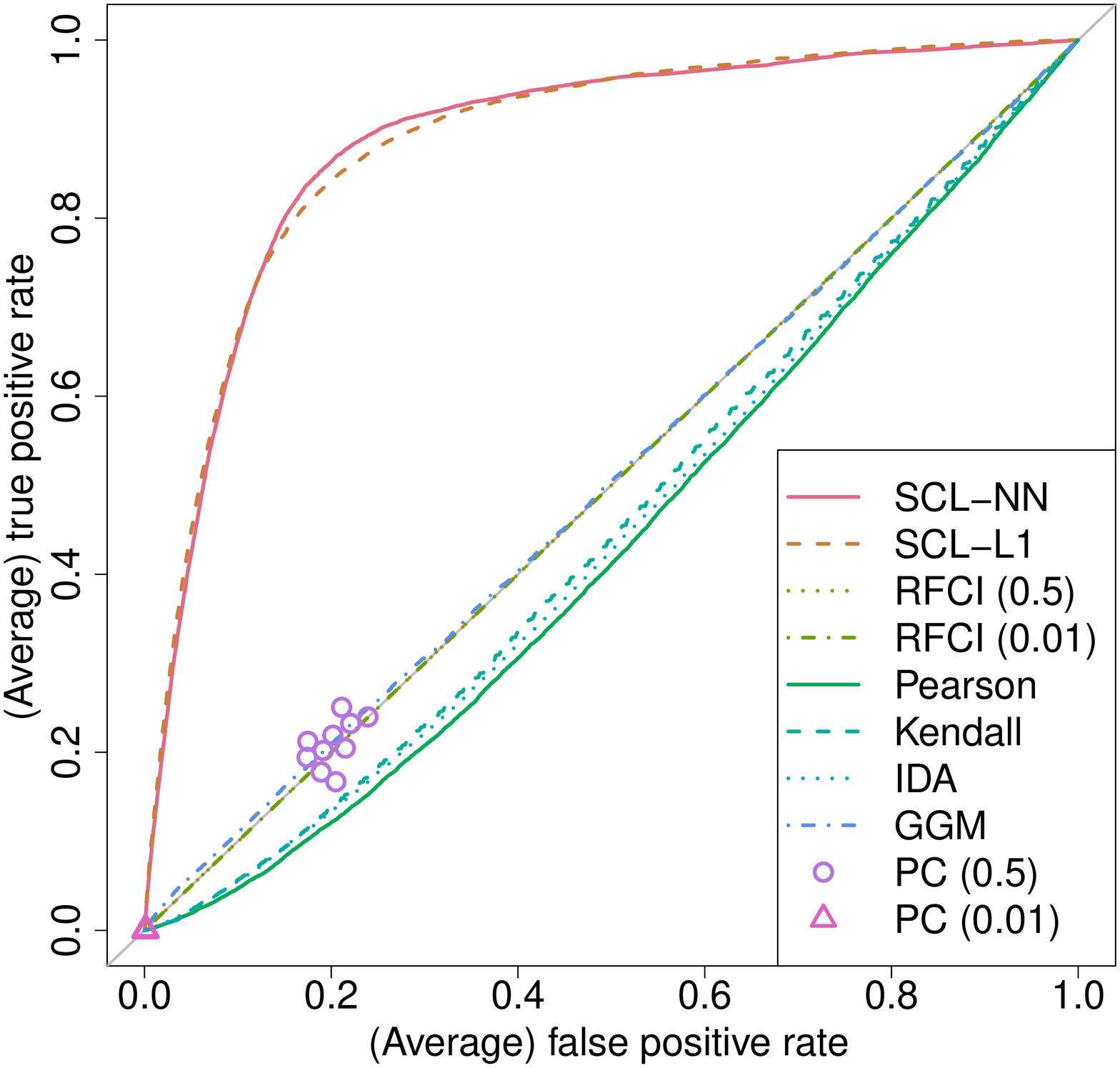}
    }
    \subcaptionbox{$p \sim 19{,}000$}{
        \includegraphics[width=0.3\textwidth]{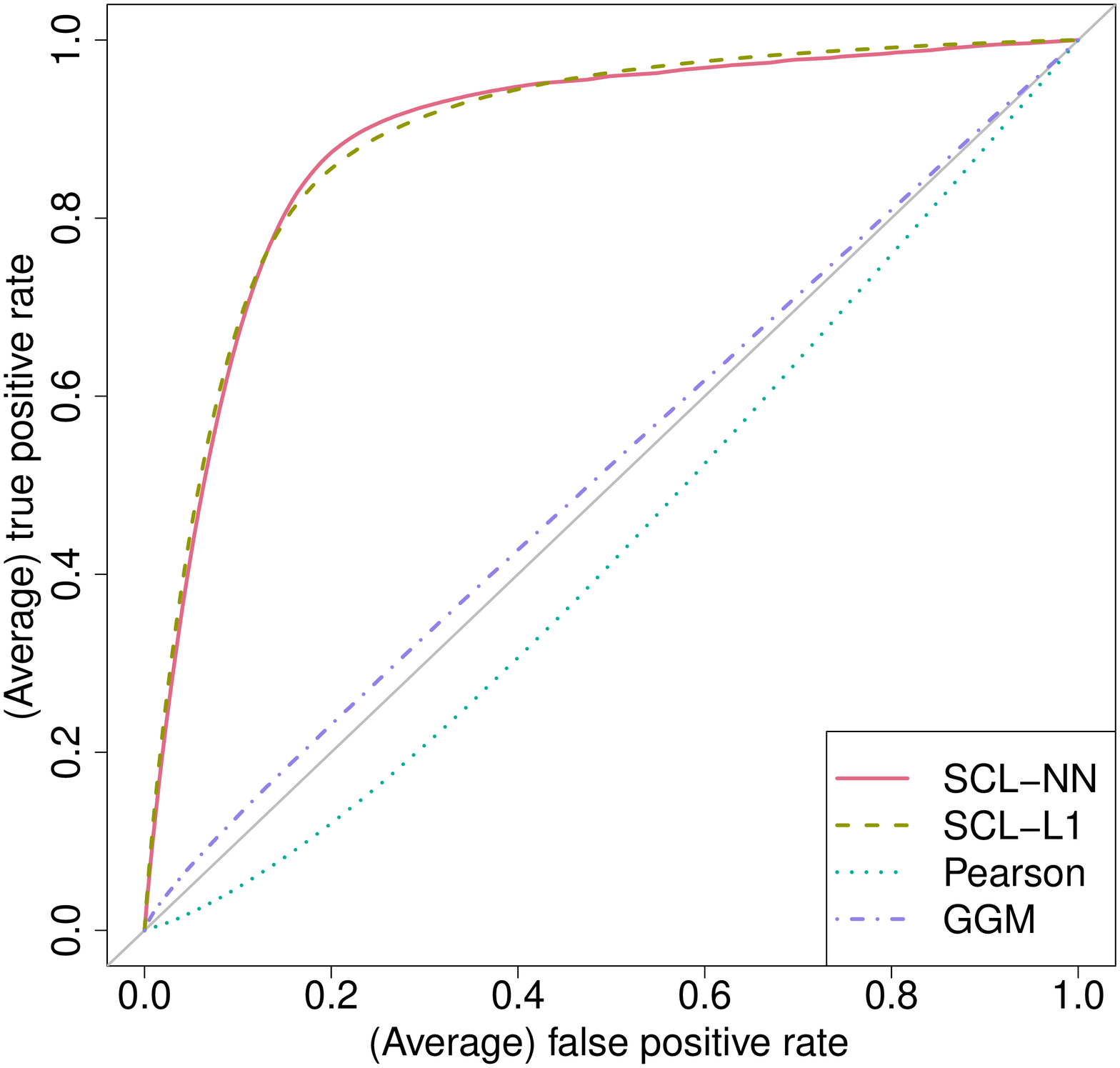}
    }
    \caption{\textit{Varying dimension $p$, random sampling}. (a) AUC vs $p$ [mean over 10 sampling iterations with error bars indicating one standard error].
    (b-d) Average ROC curves for selected regimes from (a), with locations on the ROC plane 
    (representing true positive and false positive rates)
    indicated for methods that return a point estimate of a graphical object. Causal effects were obtained from RFCI using the LV-IDA algorithm.
 }
    \label{fig:vary-p-entries}
\end{figure}

\noindent
{\it Random sampling.}
Figure~\ref{fig:vary-p-entries}(a) shows performance as a function of problem dimension $p$, with the fraction of background information $\rho$ set to 0.5 throughout (in terms of the set notation above, $\rho$ equals $|\mathcal{T}| / | \mathcal{K} |$, i.e.\ the fraction of all pairs whose causal status is known to the learner as background information; we consider varying $\rho$ below).
Figures~\ref{fig:vary-p-entries}(b-d) show the ROC curves for three values of $p$.
Results from PC (GIES is only appropriate for an ``intervention-wise'' problem set-up, see below) are shown as points on the ROC plane. For PC and RFCI/LV-IDA we show results for two different significance levels $\alpha \in \{0.01, 0.5\}$; additional results for different $\alpha$'s appear in the Appendix (Figure~\ref{fig:app-vary-alpha}).

The proposed methods perform well in absolute terms and relative to other approaches. Note that the performance initially {\it improves} with increasing dimension $p$, in line with the arguments presented above with respect to the relationship between dimension and the learning task. Computational cost is moderate even for the largest $p$ (see Figure~\ref{fig:vary-p-time} in the Appendix for wall clock times).

\bigskip

\begin{figure}
    \centering
    \subcaptionbox{}{
        \hspace{1.2cm}\includegraphics[width=0.7\textwidth]{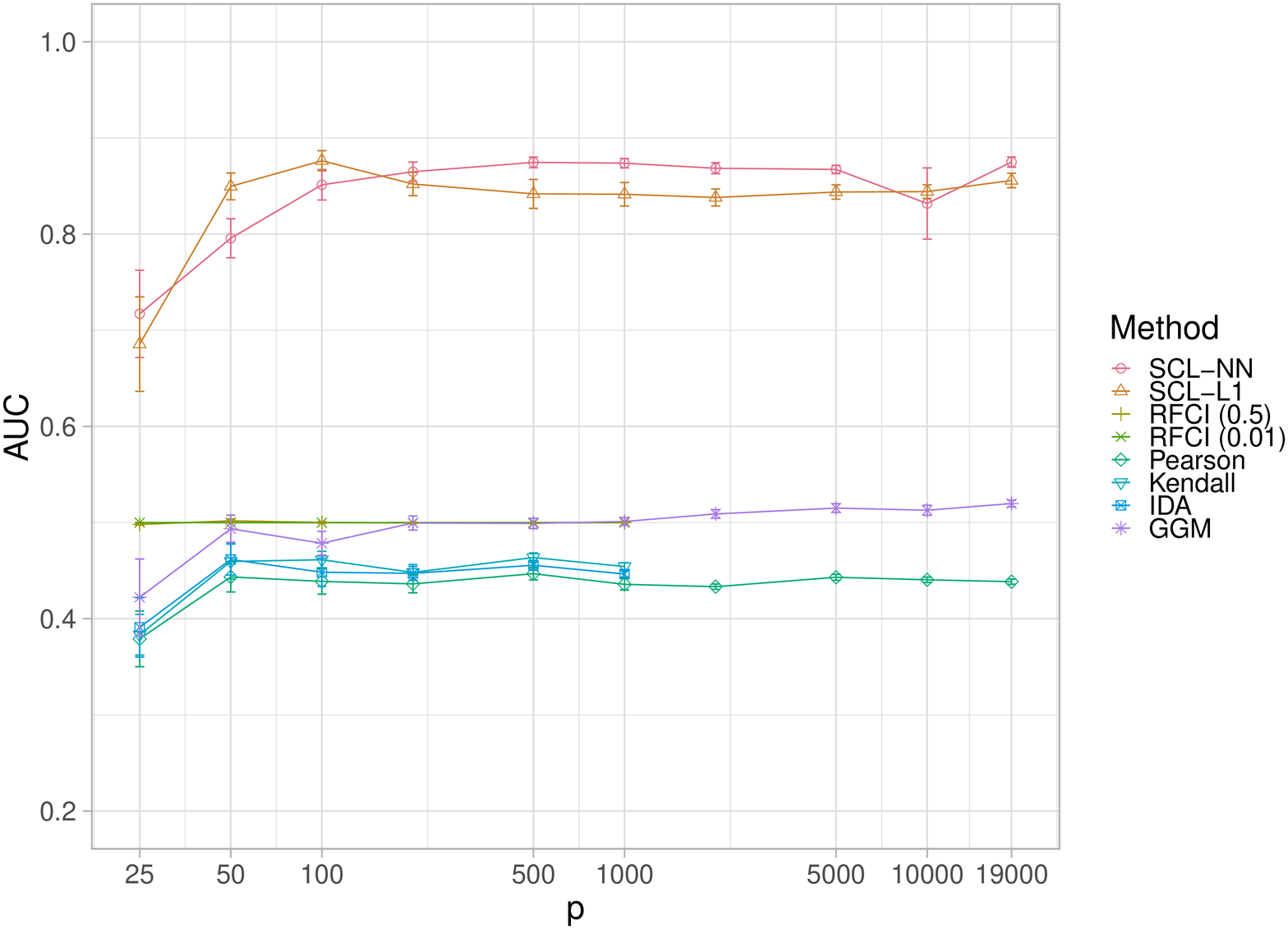}
    }
    \\
    \subcaptionbox{$p = 100$}{
        \includegraphics[width=0.3\textwidth]{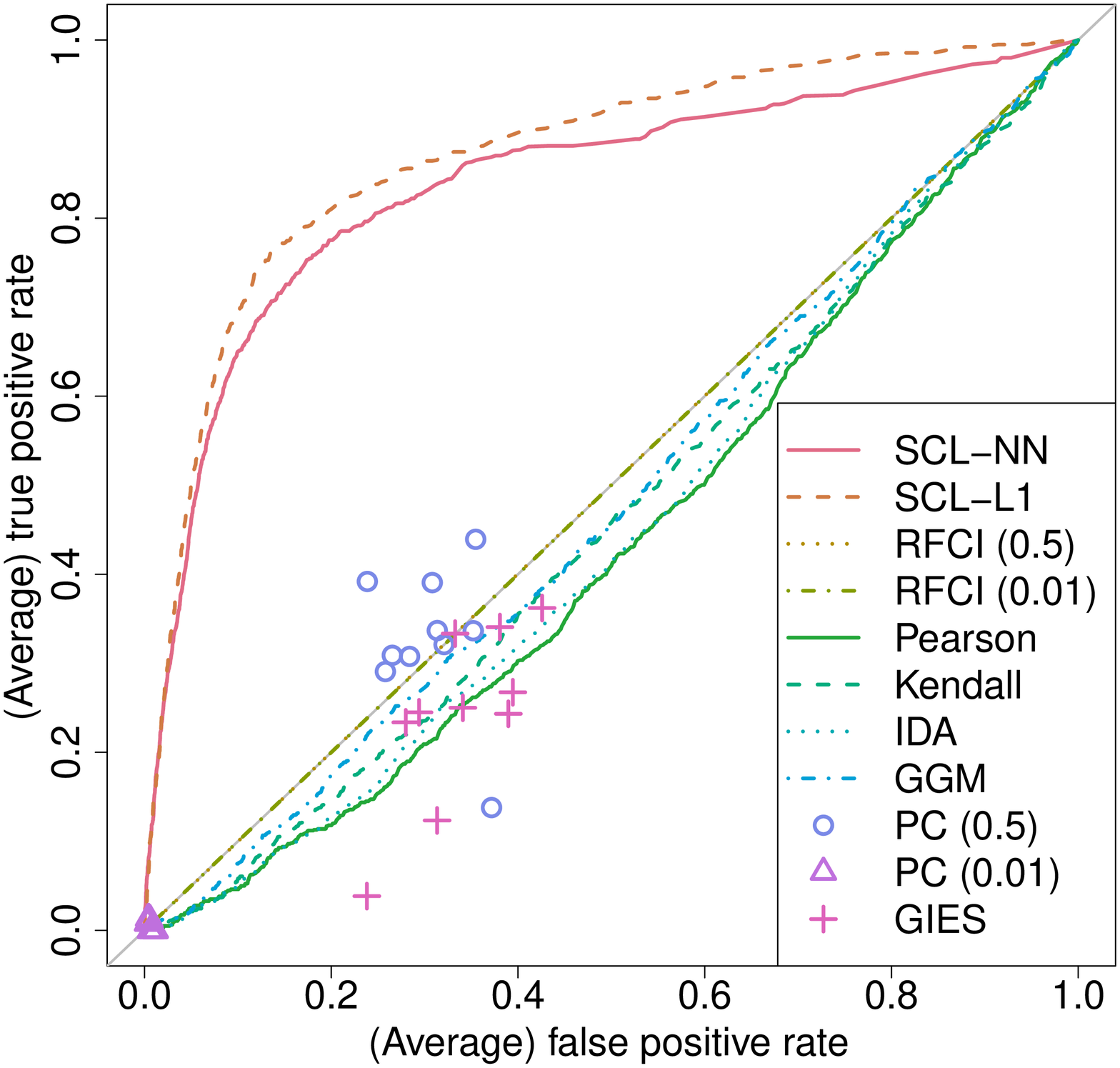}
    }
    \subcaptionbox{$p = 1000$}{
        \includegraphics[width=0.3\textwidth]{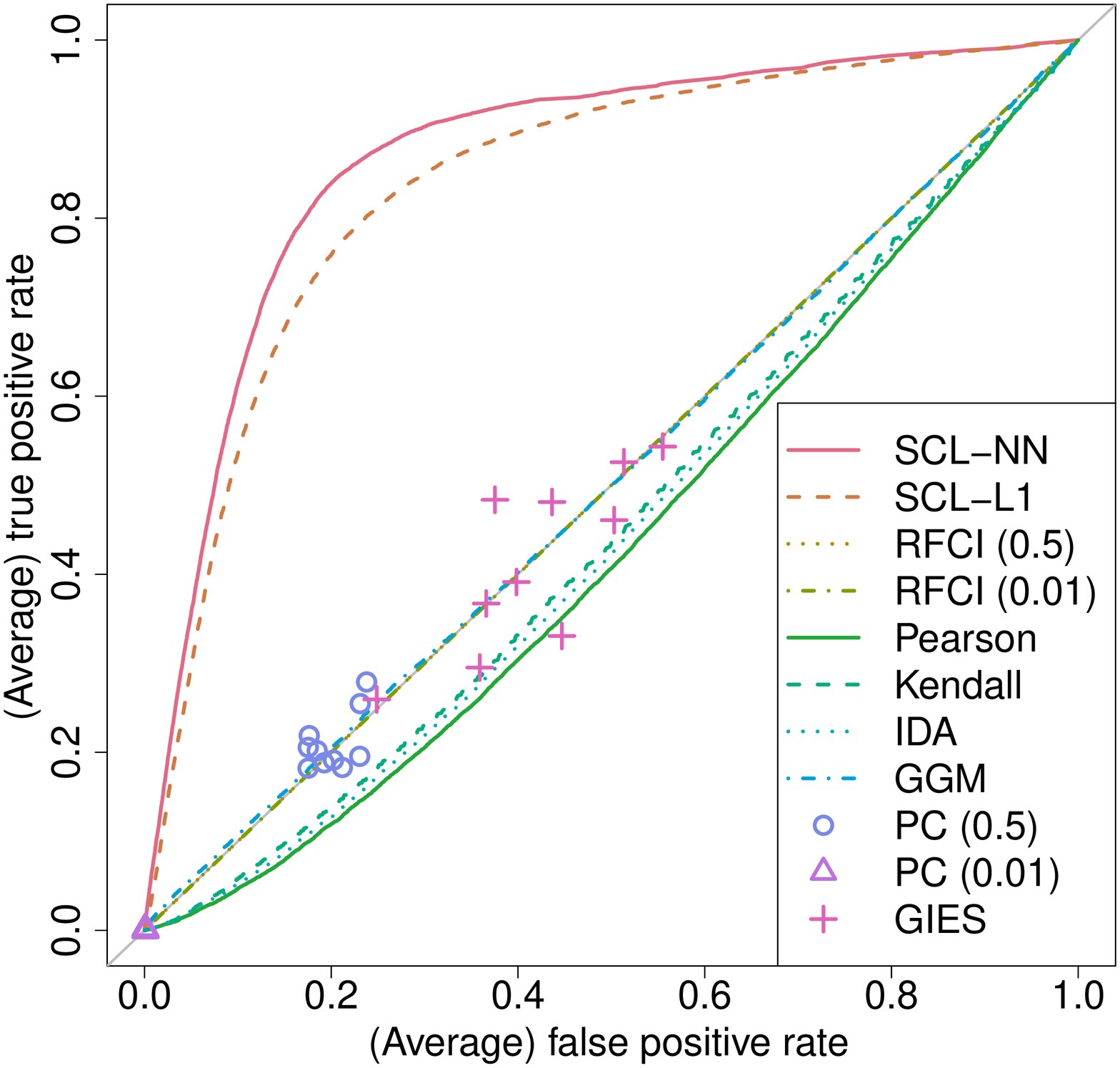}
    }
    \subcaptionbox{$p \sim 19{,}000$}{
        \includegraphics[width=0.3\textwidth]{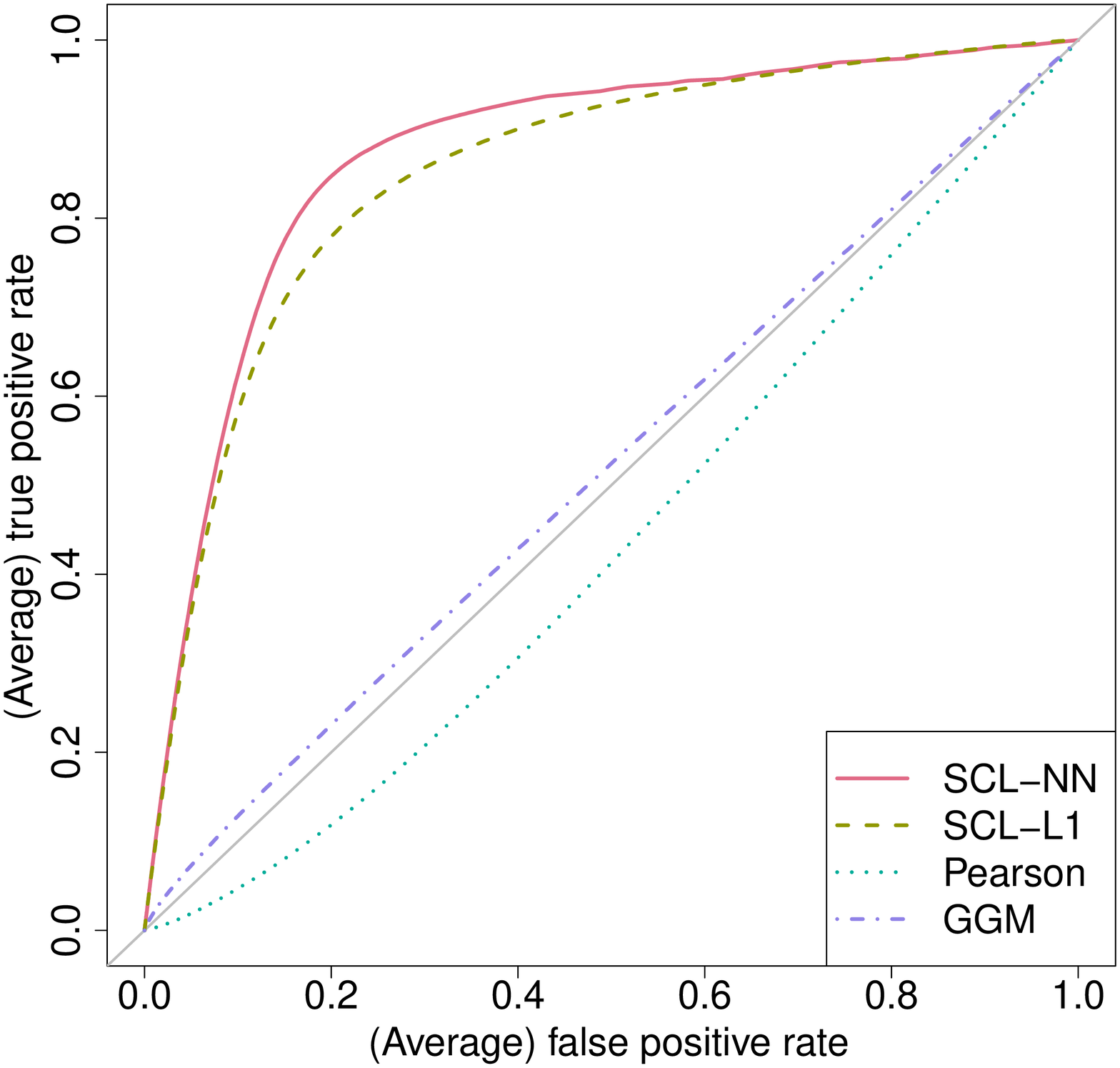}
    }
    \caption{\textit{Varying dimension $p$, intervention-wise sampling}. (a) AUC vs $p$ [mean over 10 sampling iterations with error bars indicating one standard error].
    (b-d) Average ROC curves for selected regimes from (a), with locations on the ROC plane indicated for methods that return a point estimate of a graphical object (representing true positive and false positive rates). Causal effects were obtained from RFCI using the LV-IDA algorithm.}
    \label{fig:vary-p-rows}
\end{figure}

\noindent
{\it Intervention-wise sampling.}
The results above are based on {\it random sampling} of pairs (in the sense that $\mathcal{T}$ is a random subset of $\mathcal{K}$). Figure~\ref{fig:vary-p-rows}(a) shows the corresponding result for {\it intervention-wise sampling}. The latter corresponds to the case where the learner is asked to generalize to an 
unseen intervention (and not just an unseen causal relationship or pair) and is in a sense a harder problem.
Suppose $I$ is the initial set of available interventions. Intervention-wise sampling is done by sampling a random subset $I' \subset I$ of the interventions and setting $\mathcal{T} = \{ k : i(k) \in I' \}$. This ensures that for any test pair $q \in \mathcal{Q}$, it always holds that $i(q) \notin I'$ or equivalently $q \in \mathcal{Q} \implies \nexists j : k(i(q),j) \in \mathcal{T}$.
This amounts to training on some interventions and testing on interventions that are entirely unseen in the sense that for any test pair $(i,j)$, the training pairs $\mathcal{T}$ do not include the effect of intervening on $i$ on {\it any} other variable. 

We note that GIES requires the interventional data itself, whereas our approach requires only information on known ancestral relationships (encoded in $\Pi$). For these experiments, to enable a reasonable comparison, we  provided GIES the interventional data upon which the information in $\Pi$ is based, i.e.\ precisely the interventions $I'$. Hence, in this example, GIES receives additional data (the interventional data for interventions $I'$) that the other methods do not. We note also that since GIES requires the interventional data on all variables, it cannot be obviously applied to the random sampling case above.

\subsection{Varying amount of available background information}
\begin{figure}
    \centering
    \includegraphics[width=0.94\textwidth]{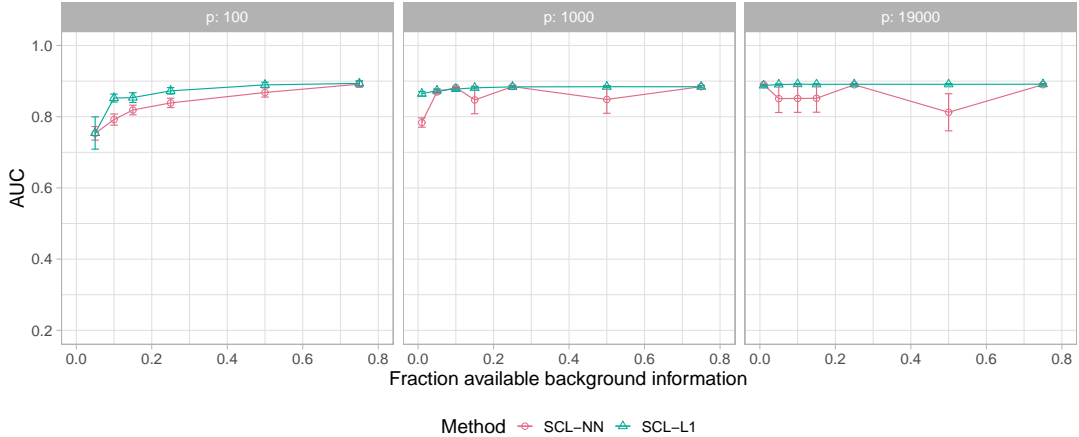}
    \caption{
        \textit{Varying fraction $\rho$ of available background information, random sampling.}  AUC shown vs fraction $\rho$ of pairs on which background information is available. Results show the mean over 10 sampling iterations with error bars indicating one standard error.
    }
    \label{fig:vary-percentage-entries}
\end{figure}

\begin{figure}
    \centering
    \subcaptionbox{}{
        \includegraphics[width=0.94\textwidth]{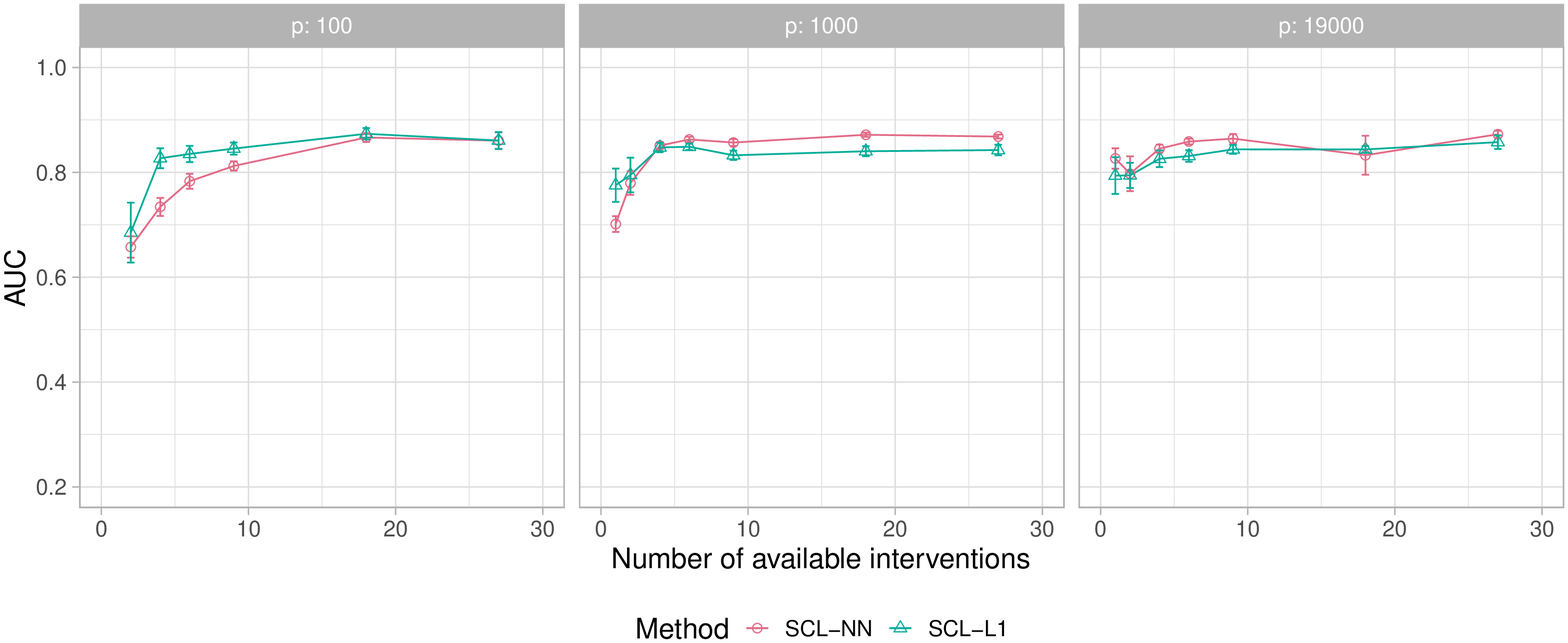}
    }
    \\
    \vspace{0.2cm}
    \hdashrule[0.5ex]{0.9\textwidth}{0.25pt}{2pt}\\
    \vspace{0.2cm}
    $p = 100$
    \\
    \subcaptionbox{$\rho = 0.25$}{
        \includegraphics[width=0.25\textwidth]{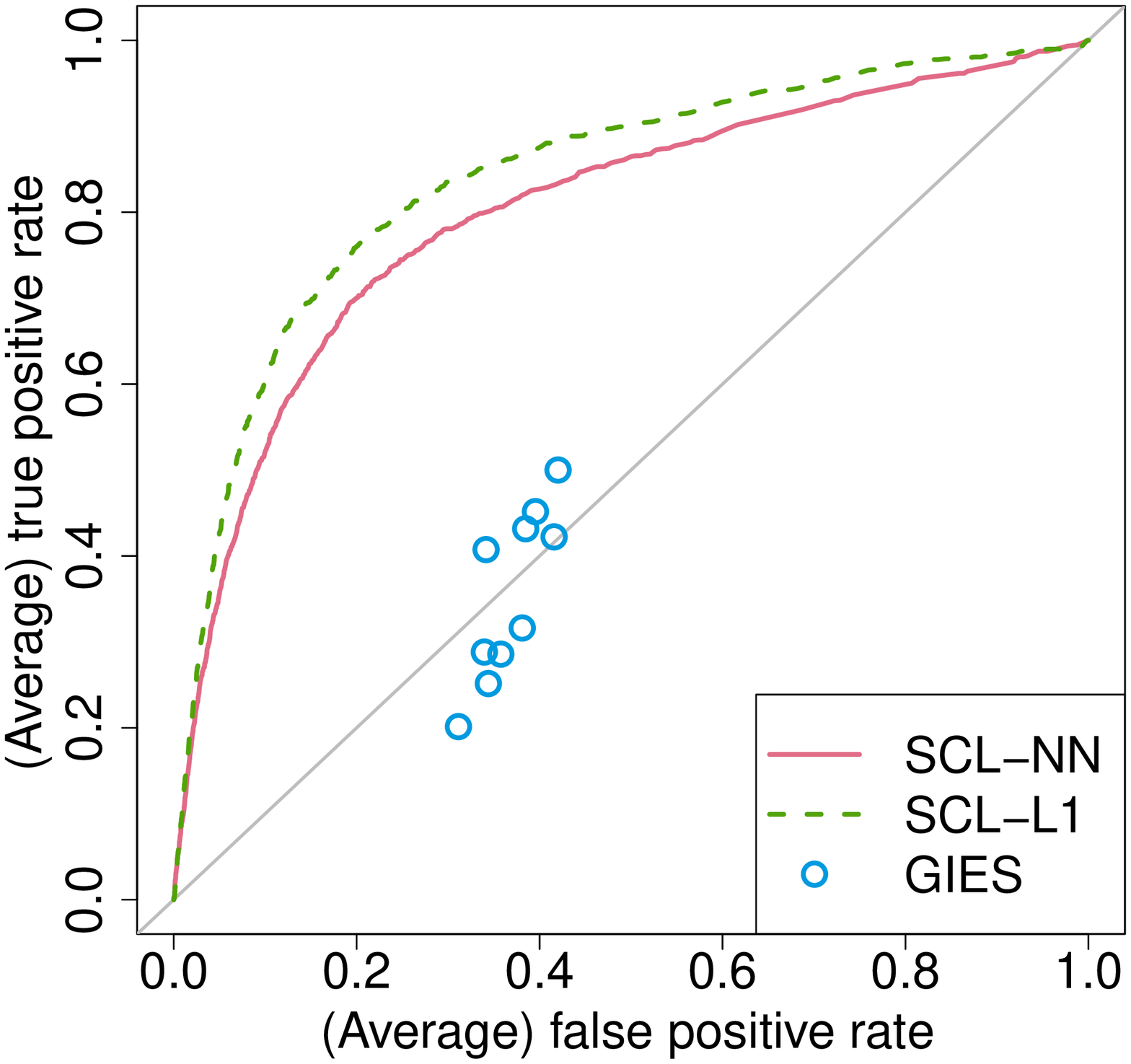}
    }
    \subcaptionbox{$\rho = 0.5$}{
        \includegraphics[width=0.25\textwidth]{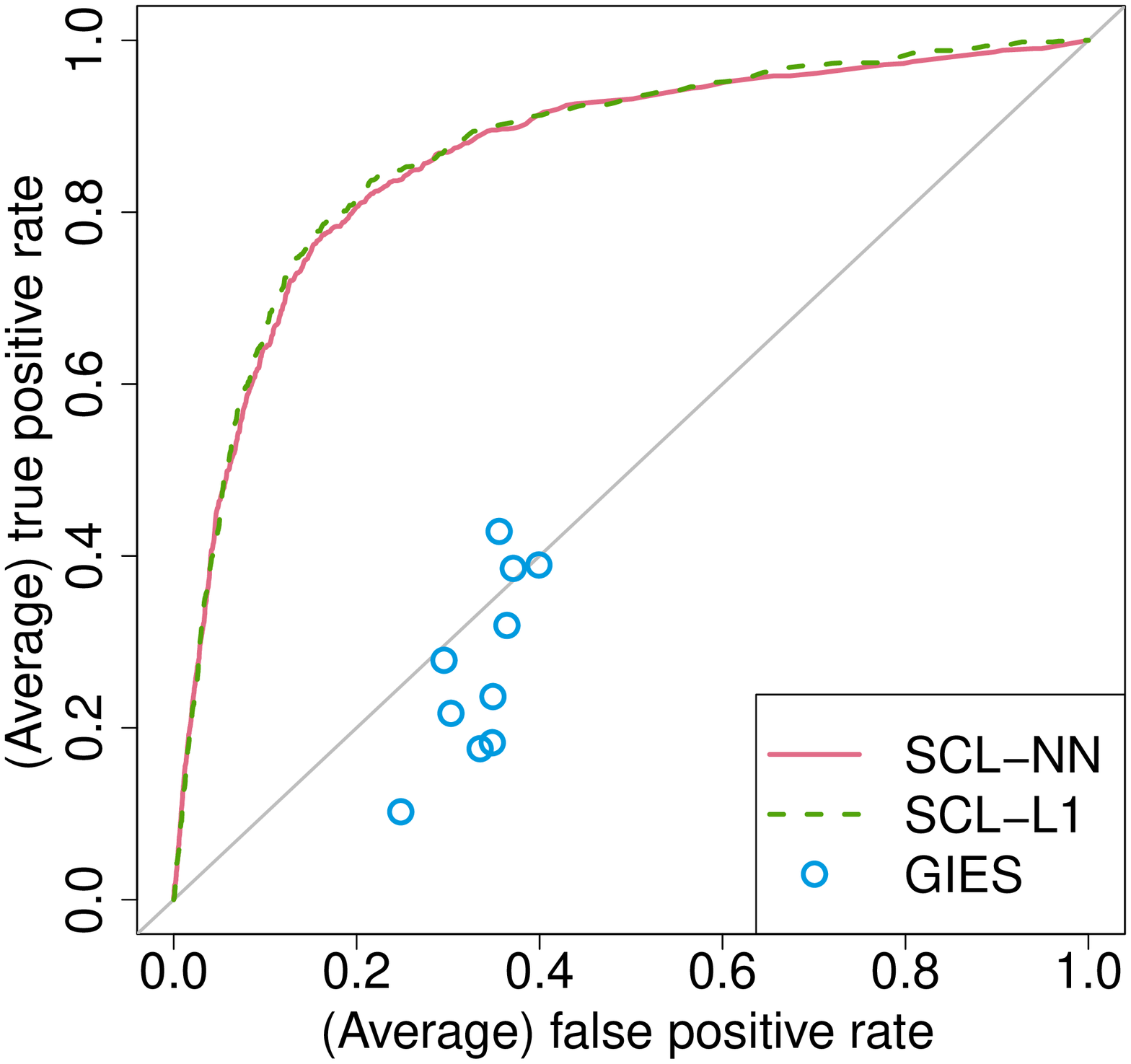}
    }
    \subcaptionbox{$\rho = 0.75$}{
        \includegraphics[width=0.25\textwidth]{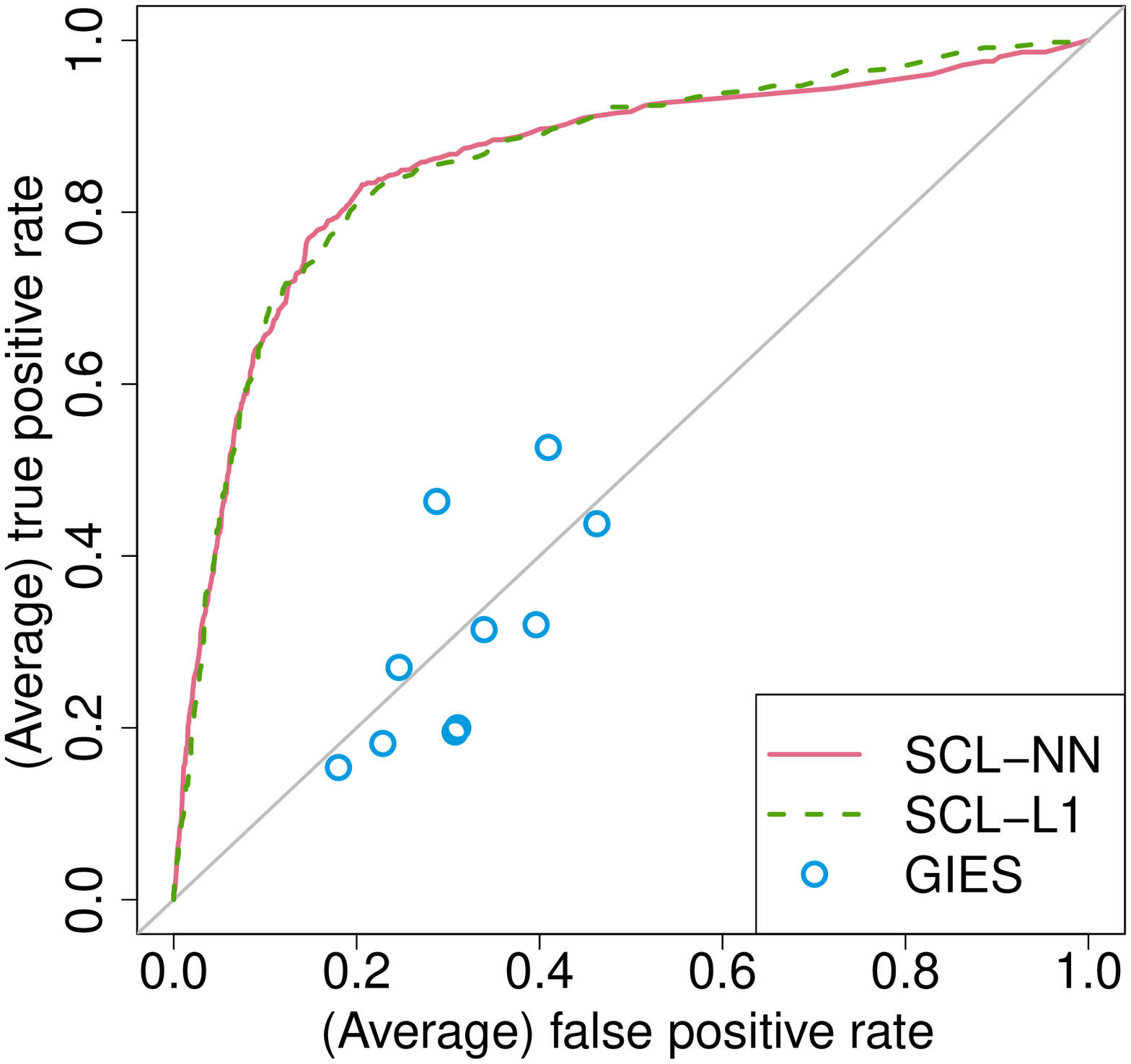}
    }
    \\
    \vspace{0.2cm}
    \hdashrule[0.5ex]{0.9\textwidth}{0.25pt}{2pt}\\
    \vspace{0.2cm}
    $p = 1000$
    \\
    \subcaptionbox{$\rho = 0.25$}{
        \includegraphics[width=0.25\textwidth]{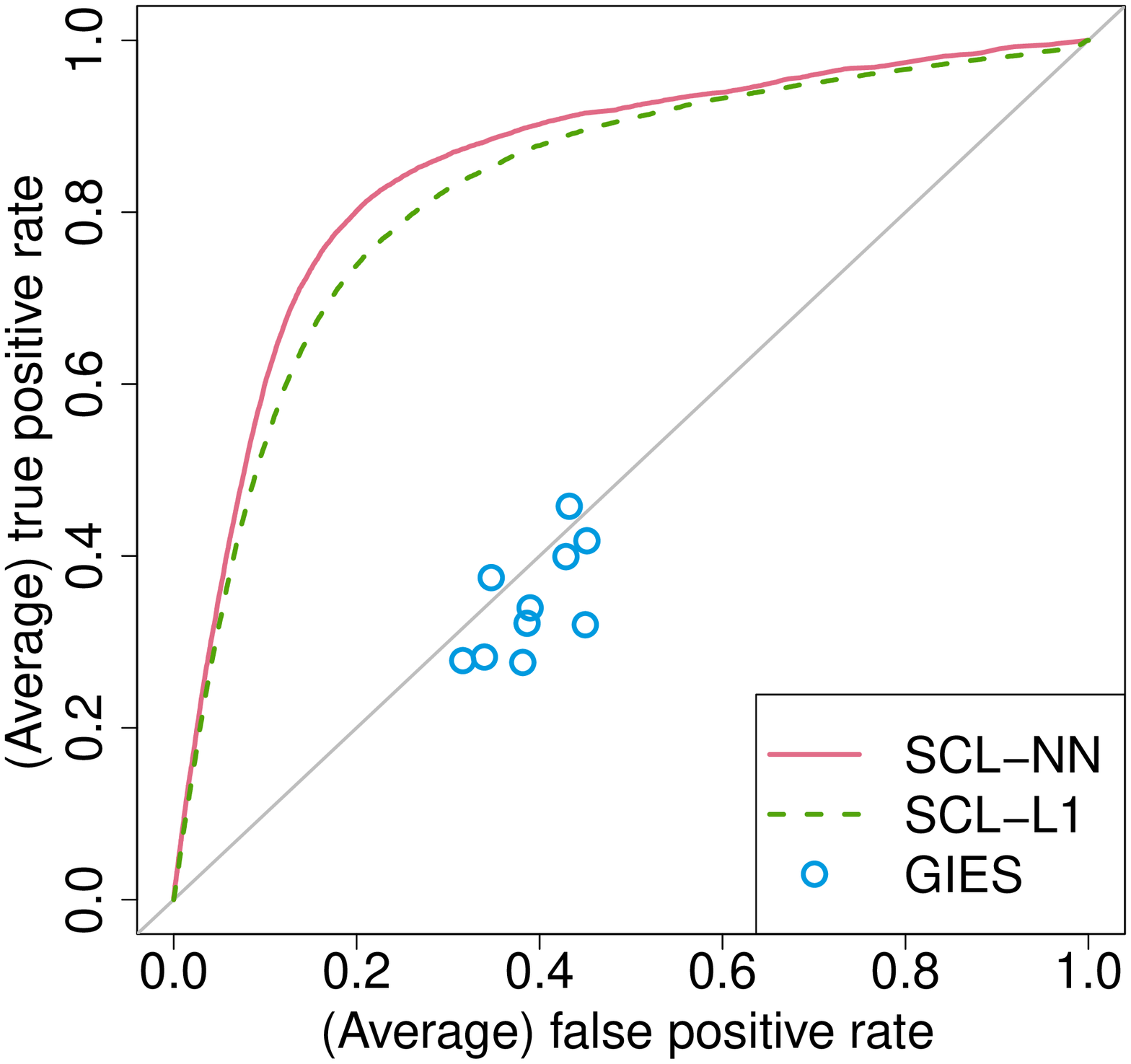}
    }
    \subcaptionbox{$\rho = 0.5$}{
        \includegraphics[width=0.25\textwidth]{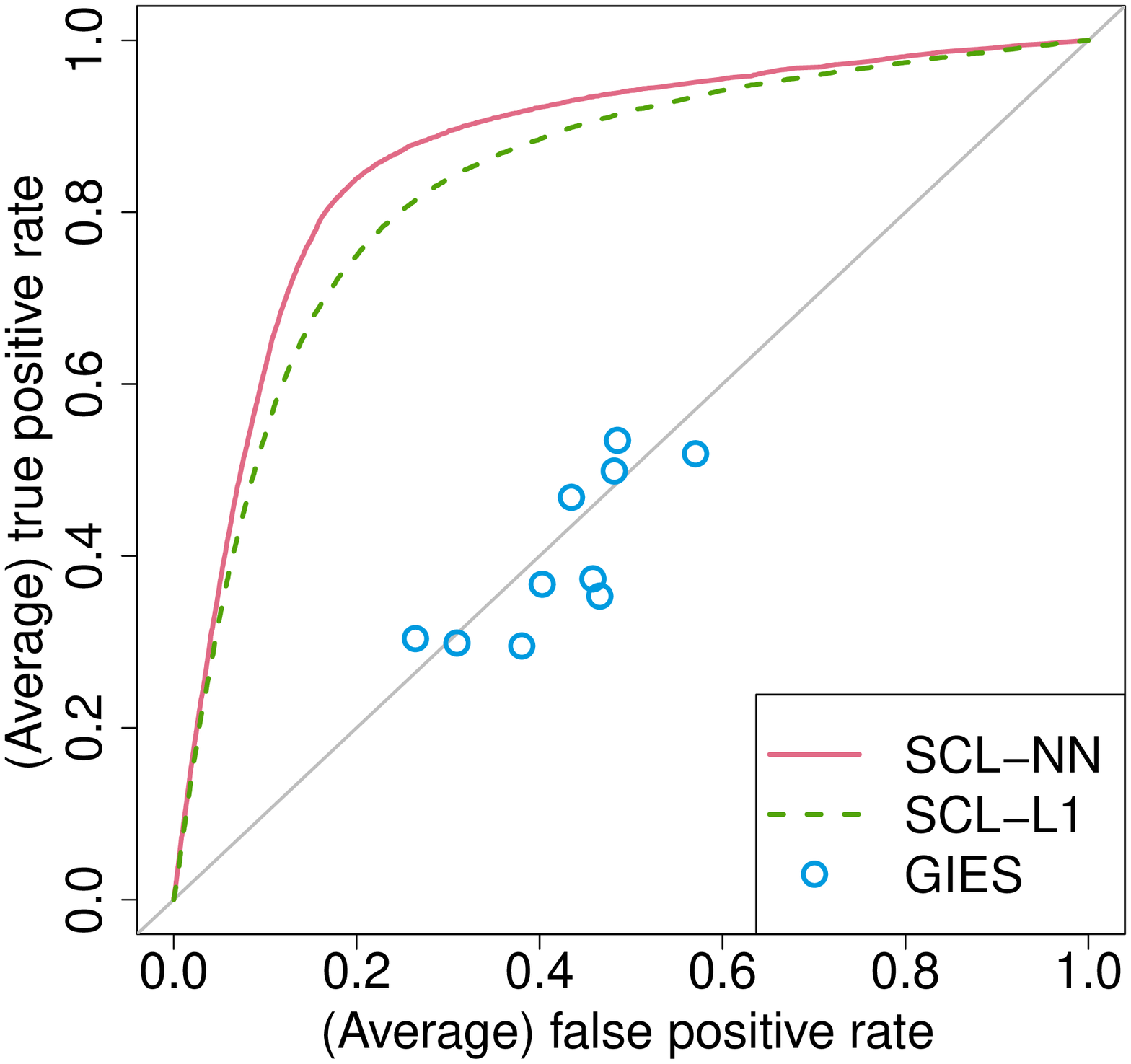}
    }
    \subcaptionbox{$\rho = 0.75$}{
        \includegraphics[width=0.25\textwidth]{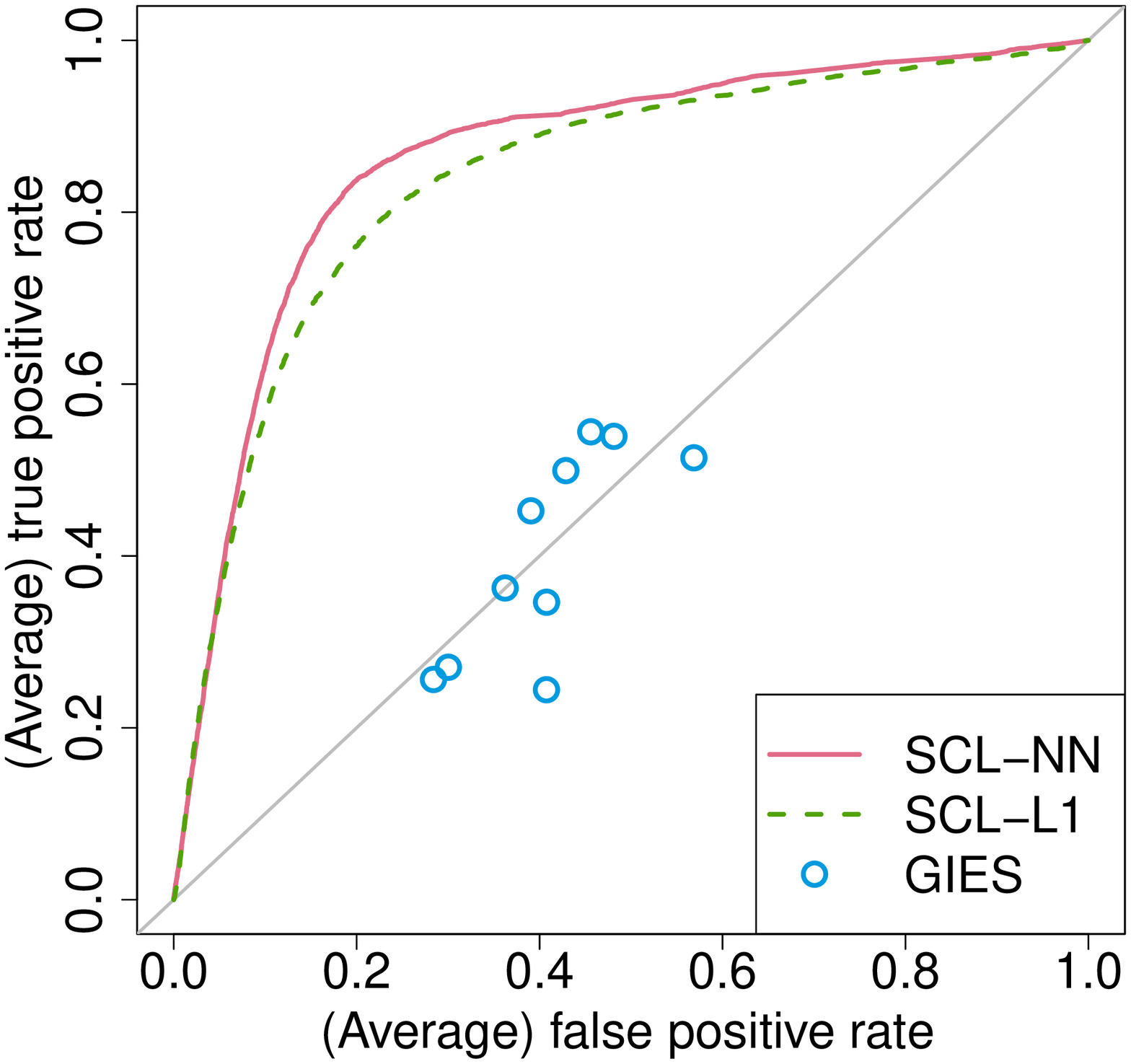}
    }
    \vspace{0.4cm}
    \caption{
        \textit{Varying amount of available background information, intervention-wise sampling.} 
        (a) AUC vs amount of background information (absolute number of available interventions indicated on the horizontal axis) [mean over 10 sampling iterations with error bars indicating one standard error].
        (b-g) Average ROC curves for selected regimes for (b-d) $p=100$ and (e-g) $p=1000$, with locations on the ROC plane indicated for GIES (for each of the 10 iterations).
    }
    \label{fig:vary-percentage-rows}
\end{figure}

The previous results are based on a fixed fraction $\rho$ of pairs on which background information is available. Next, we considered varying the fraction $\rho$. This is practically relevant since in many settings information may only be available on a relatively small fraction of pairs.
Figure~\ref{fig:vary-percentage-entries} 
shows performance as a function of $\rho$ for the random sampling case. 
Figure~\ref{fig:vary-percentage-rows}(a) shows the corresponding result for the intervention-wise case. In both cases, we find that performance improves rapidly with the fraction $\rho$. Since only GIES from among the other methods varies with $\rho$, 
we show only results for the proposed approaches and GIES (in the intervention-wise case). 
We were not able to run GIES for the very high dimensional examples due to computational demands. 
For the other methods, performance would be as shown in the previous figures, since for these methods $\rho$ plays no role.

\subsection{Sensitivity to  background information}
\label{sec:sensitivity}
The learner relies on knowledge $\Pi$ of some ancestral causal relationships. In this section we consider sensitivity to perturbation of such information. The intention is  to investigate the effect  of including in $\Pi$ information that is in fact {\it incorrect}. 
This was done by intentionally perturbing the input labels in such a way as to control the fraction of labels altered whilst keeping the overall sparsity (i.e.\ the fraction of ancestral causal edges in the full graph) fixed. 

To clarify how this was done, we introduce some additional notation. In what follows let $\tilde{\Pi}$ denote a perturbed  version of the background knowledge. For the experiments all learning was done as described above, but with the perturbed $\tilde{\Pi}$ in place of $\Pi$.
Denote by $\mathcal{T}^{(1)} = \{ k : k \in \mathcal{T} \wedge y_k(\Pi) = 1 \}$ the set of all causal pairs known from background knowledge $\Pi$ and by $\mathcal{T}^{(0)} = \mathcal{T} \setminus  \mathcal{T}^{(1)}$ the complement (i.e.\ pairs known to be non-causal from background knowledge). 
Let $\tilde{\mathcal{T}}^{(1)} \subset \mathcal{T}^{(1)}$ be a random subset of the positive cases and similarly $\tilde{\mathcal{T}}^{(0)} \subset \mathcal{T}^{(0)}$ for the negatives (with $|\tilde{\mathcal{T}}^{(0)}| = | \tilde{\mathcal{T}}^{(1)}|$) and $\tilde{\mathcal{T}} = \tilde{\mathcal{T}}^{(0)} \cup \tilde{\mathcal{T}}^{(1)}$ be the complete set of perturbed pairs. Now, we perturb the labels as $$\tilde{y}_k =
\begin{cases}
    0 & \text{if } k \in \tilde{\mathcal{T}}^{(1)} \\
    1 & \text{if } k \in \tilde{\mathcal{T}}^{(0)} \\
    y_k & \text{otherwise}
\end{cases}
$$

This means that exactly $|\tilde{\mathcal{T}}^{(1)}|$ of the known causal pairs are perturbed whilst leaving the total number of ``known" casual pairs as input to the learner unchanged. 
The above procedure guarantees that (i) causal sparsity is unchanged in the sense that 
$\frac{1}{|\mathcal{T}|} \sum_{k \in \mathcal{T}} y_k = \frac{1}{|\mathcal{T}|} \sum_{k \in \mathcal{T}} \tilde{y}_k$, and (ii) input information on the perturbed pairs  is indeed incorrect, i.e.\ \mbox{$\forall k \in \tilde{\mathcal{T}} : \tilde{y}_k \neq y_k$}.

\begin{figure}[tb]
    \centering
    \includegraphics[width = 0.8\textwidth]{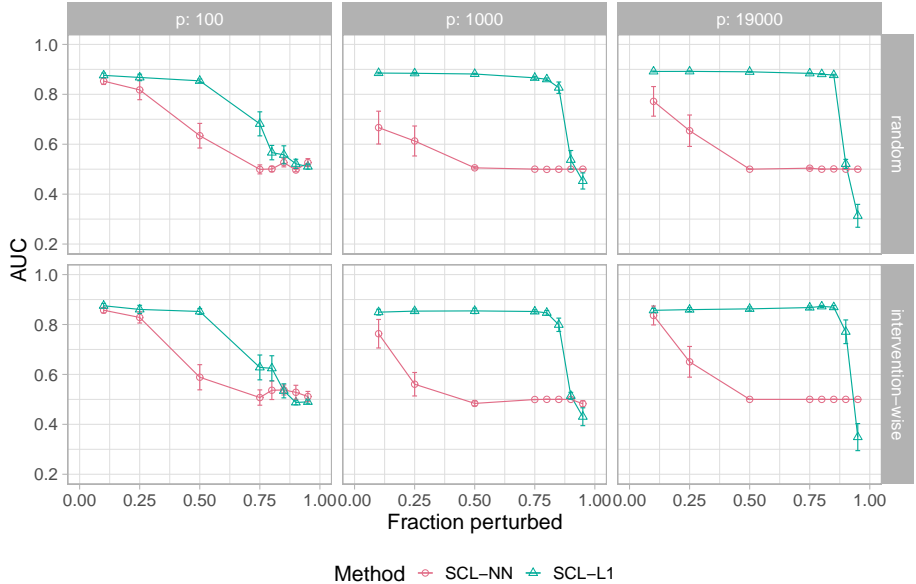}
    \caption{\textit{Sensitivity to perturbation of background information}. Here, $\Pi$ was perturbed by introducing corrupted information. Performance is shown as a function of the fraction of background information that was perturbed (see text for details). 
    Upper and lower panels are for the random sampling and intervention-wise sampling cases respectively.
    Results show the mean over 10 sampling iterations with error bars indicating one standard error.}
    \label{fig:swap-entries}
\end{figure}

Figure~\ref{fig:swap-entries} shows performance as a function of the fraction $|\tilde{\mathcal{T}}^{(1)}|/|\mathcal{T}^{(1)}|$ 
of the input causal information that was corrupted as described above. Performance is highly robust for the $\ell_1$ approach but the neural network is dramatically more fragile. We note that this is probably not a fundamental difference between neural networks and sparse regression and could likely be rescued with a different architecture and/or regularization strategy. Rather, the point is that a highly flexible model can easily fit the input information, hence for corrupted input information the simplicity of the $\ell_1$ model is an advantage and amounts to stronger regularization. 

\subsection{Correcting background information}

\begin{figure}[tb]
    \centering
    \includegraphics[width = 0.8\textwidth]{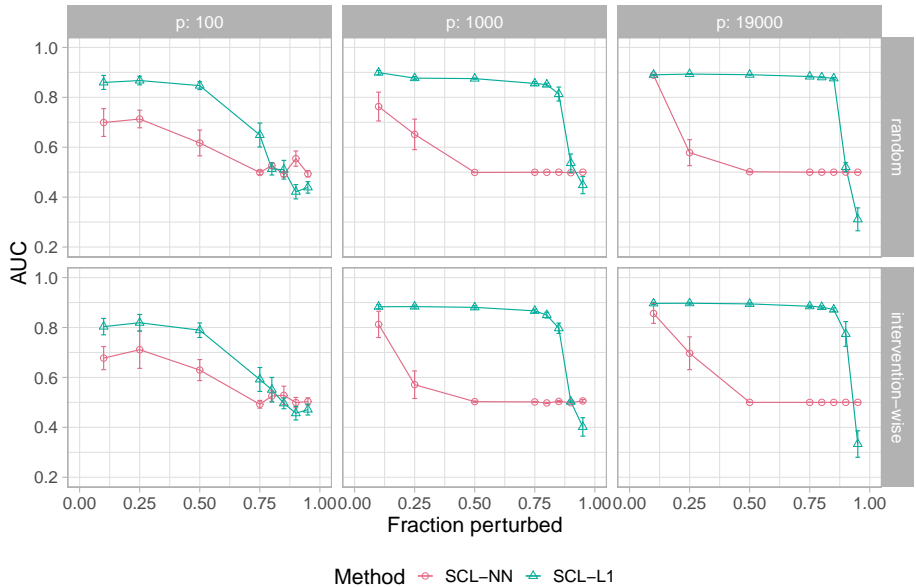}
    \caption{\textit{Error correction of incorrect prior information.} Learning was carried out using incorrect background information. The models were used to correct the input information and assessment was done with respect to the true (i.e.\ pre-perturbation) input information. AUC is shown as function of the fraction of background information that was perturbed (see text for details). Upper and lower panels are for the random sampling and intervention-wise sampling cases respectively. Results show the mean over 10 sampling iterations with error bars indicating one standard error.}
    \label{fig:relabelling}
\end{figure}

As described in Section \ref{sec:correction}, the learner could in principle be used to {\it correct}  prior information. Here, we investigate the behavior of this approach in the perturbed input setting described in the previous Section, taking advantage of the fact that we have intentionally induced errors.
Figure~\ref{fig:relabelling} shows performance with respect to error correction, as a function of the fraction $|\tilde{\mathcal{T}}^{(1)}|/|\mathcal{T}^{(1)}|$ of the input causal information that was corrupted. Note that here performance is assessed  with respect to the set of perturbed pairs $\tilde{\mathcal{T}}$. 
Although these are a subset of the {\it training} data,  the inputs are by design incorrect. Hence, in the present case an AUC of unity implies that the learner corrects the input labels to their true (i.e.\ pre-perturbation) values. We find that the $\ell_1$ learner is highly effective at error correction in the high dimensional settings: this is a blessing of dimensionality and is due to the fact that in larger problems there is more information and therefore more opportunity to detect regularities that can then be used to cope with incorrect inputs and ultimately, as shown here, rectify such inputs. 

\subsection{Knowledge of a small number of ancestral relationships}

In fields such as economics, epidemiology or the social sciences, although large scale interventional experiments of the kind now performed in molecular biology are not possible, some ancestral relationships are typically known from the science. However, these tend not be the total effects of a certain intervention on a large number of variables, but rather isolated facts (corresponding in our notation to specific pairs $(i,j)$). 
To take one example: an expert epidemiologist might not be able to provide prior information on the effect of any intervention on {\it all} measured variables in a high-dimensional study, but might be able to specify some pairs (e.g.\ connecting specific exposures to specific phenotypes). Here, we take advantage of the molecular data in our case study to ask whether learning is possible if the background information $\Pi$ contains only a small number of such ancestral relationships. That is, we consider the case where $\Pi$ provides information on only  a small number of pairs $(i,j)$ where it is known that $i$ is a causal ancestor of $j$ (i.e.\ only positive information). 

\begin{figure}[tb]
    \centering
    \includegraphics[width = 0.8\textwidth]{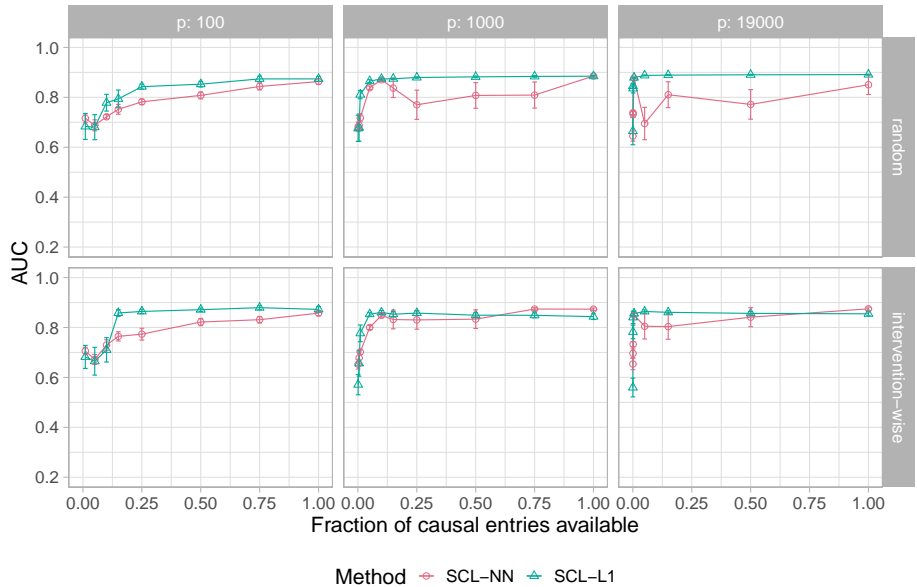}
    \caption{\textit{Learning using only a small number of known ancestral relationships.} The horizontal axis shows the fraction $|\mathcal{T}^{+}| / |\mathcal{T}^{(1)}|$ of available ancestral  relationships that were available to the learner (see text for details). Results show the mean over 10 sampling iterations with error bars indicating one standard error.}
    \label{fig:sparse-labelling}
\end{figure}

To study this case, rather than giving the learners access to background information as in the examples above, we allowed access only to information on a specified fraction of (positive) ancestral relationships. 
Specifically, let $\mathcal{T}^{+} \! \subset \! \mathcal{T}^{(1)}$ be a (possibly small) random subset of the causal pairs known from background knowledge (notation as in Section \ref{sec:sensitivity}). 
We applied the proposed methods using this information to construct the input labels $y_k$ by simply assuming {\it all other pairs} as non-causal, i.e.\ setting $\tilde{y}_k =1$ if $k \in \mathcal{T}^{+}$ and to zero otherwise. Figure~\ref{fig:sparse-labelling} shows results for the random sampling and intervention-wise cases. Performance improves rapidly with the number $| \mathcal{T}^{+} |$ of positive examples available and the gain is more rapid for higher dimensional examples. The $\ell_1$ approach is more effective (likely due to model complexity as discussed above). 
As a point of comparison, we considered the case of locating  the ``known" ancestral examples entirely at random (that is, randomly setting $\tilde{y}_k =1$). Learning failed in all cases, with AUCs approximately 0.5 (see Figure~\ref{fig:app-sparse-labelling-random} in the Appendix).


%
%
%

\section{Discussion}

In this paper, we presented a simple,  discriminative approach to causal learning that we showed can be effectively applied at human genome-wide scale. The general approach we propose is modular in the sense that a  variety of specific featurizations and learners could be used. The specific approach of SCL that we focused on in our empirical work scaled demonstrably well, both in terms of computational demand and statistical performance. 

Although  effective at the specific tasks that we considered, in contrast to graphical models-based methods, SCL  is less general in the sense that it only learns ancestral relationships, but cannot provide a full range of probabilistic output (e.g.\ post-intervention distributions). Nevertheless, its good performance and favourable scaling suggest that it may be fruitful to investigate combining SCL with graphical models-based approaches in future.

In molecular biology, the question of how to learn causally meaningful, systems-level networks  remains open. If this were possible at scale, the implications for study design would be considerable. Our results suggest that such scalable causal learning may be possible, but further empirical work will be needed in future to fully understand the scope and limitations of our approach in diverse biological contexts.

%% file: supplementary.tex
\section{Additional results}

\begin{figure}[tbh]
    \centering
    \includegraphics[width=0.55\textwidth]{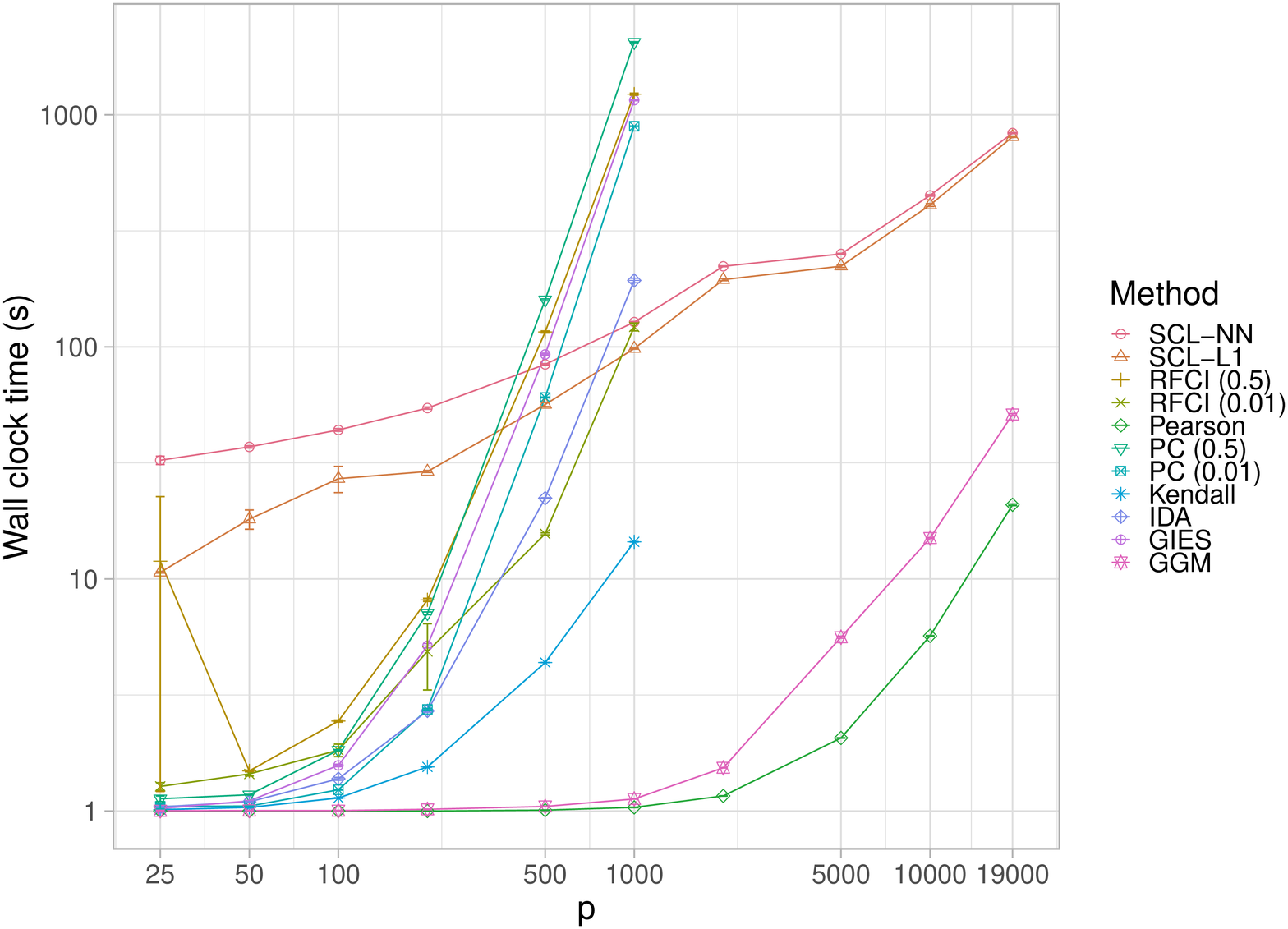}
    \caption{Wall clock times of the algorithms as a function of dimension $p$. Results show the mean over 10 sampling iterations with error bars indicating one standard error.}
    \label{fig:vary-p-time}
\end{figure}

\begin{figure}[tbh]
\captionsetup[subfigure]{labelformat=empty}
    \centering
    $p = 100$ \\
    \begin{subfigure}[b]{0.23\textwidth}
        \centering
        \includegraphics[width=\textwidth]{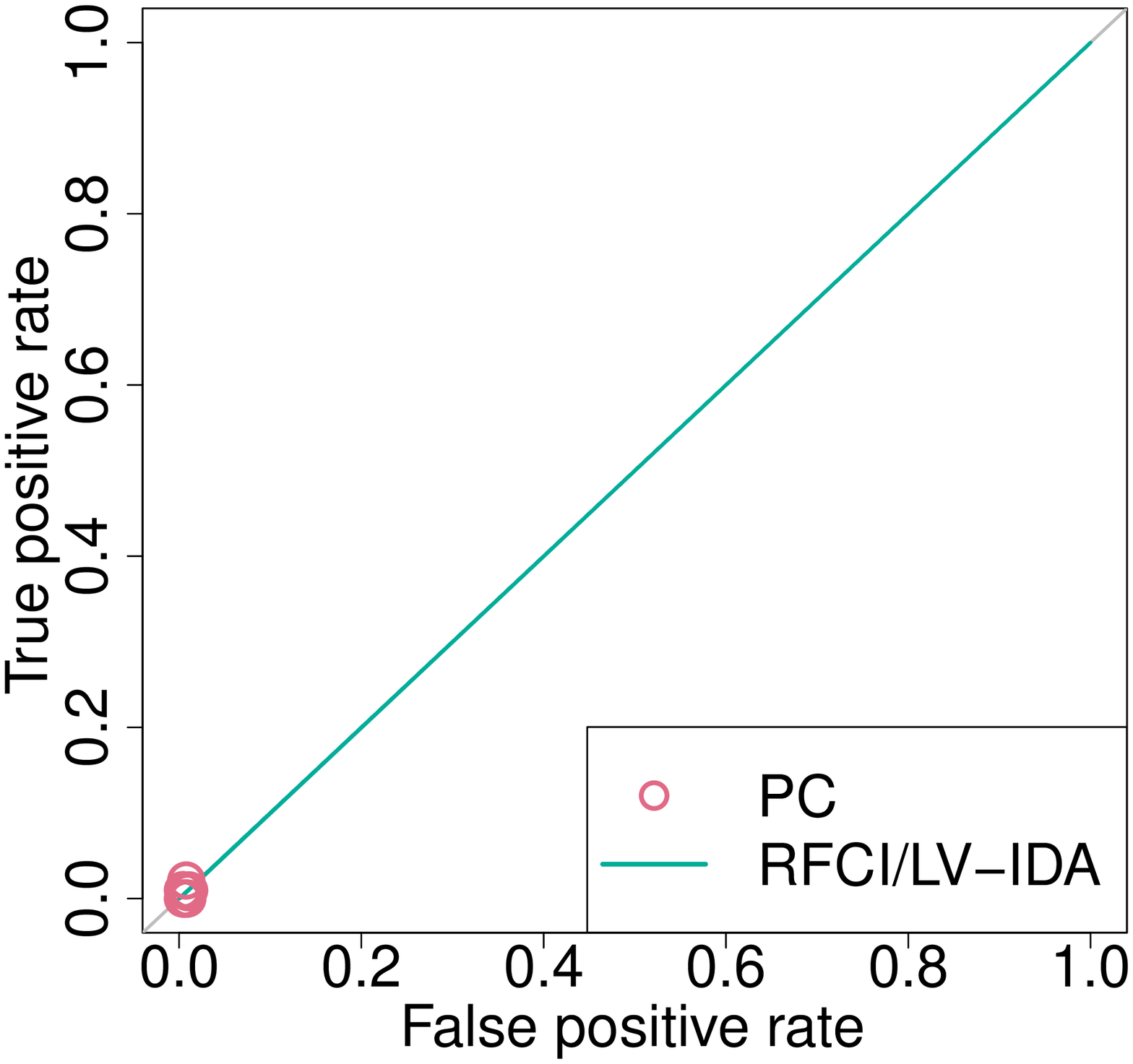}
        \caption{$\alpha = 0.01$}
        \label{fig:app-vary-alpha-1}
    \end{subfigure}
    \begin{subfigure}[b]{0.23\textwidth}
        \centering
        \vspace{-.4cm}
        \includegraphics[width=\textwidth]{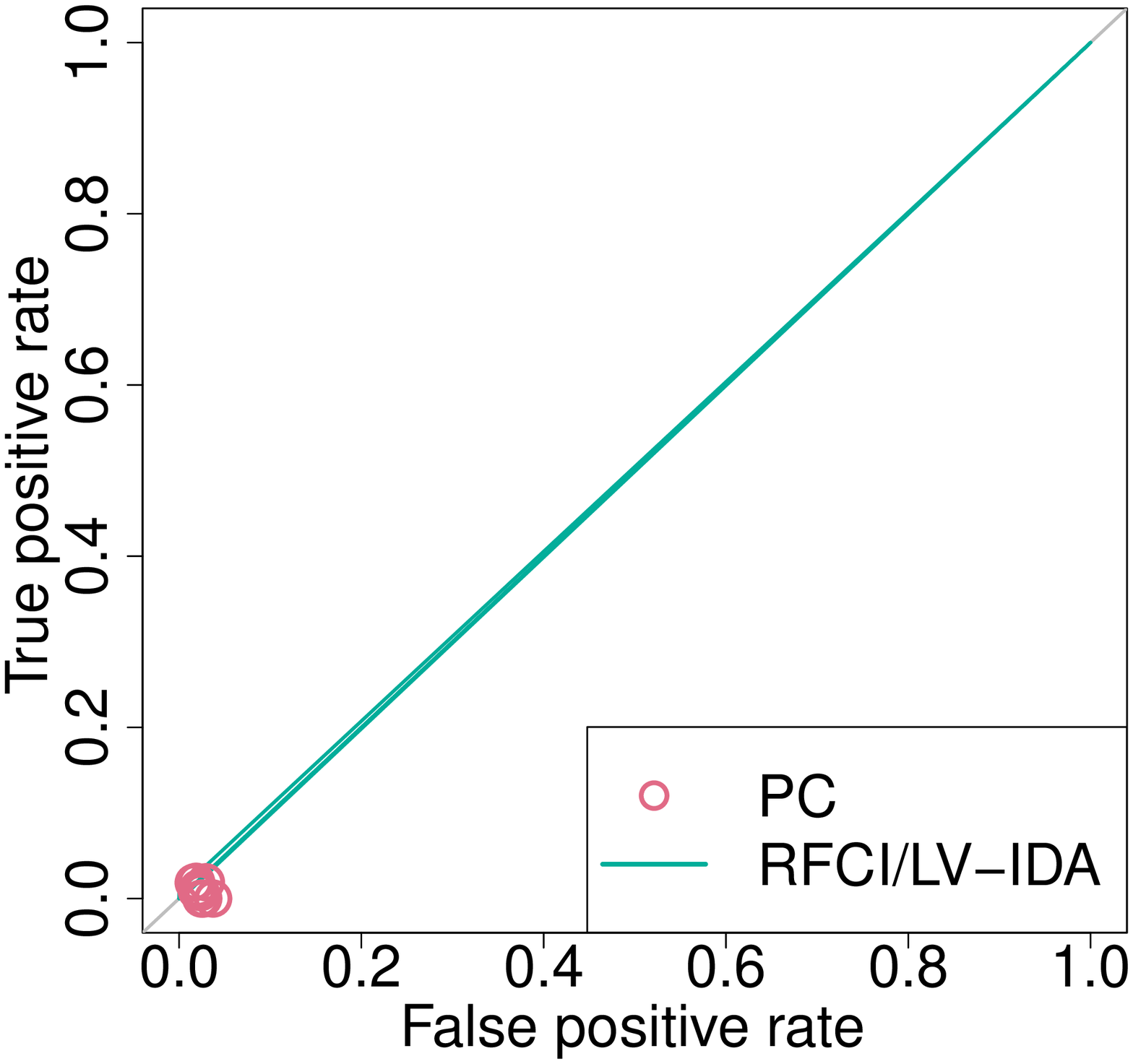}
        \caption{$\alpha = 0.05$}  
        \label{fig:app-vary-alpha-2}
    \end{subfigure}
    \begin{subfigure}[b]{0.23\textwidth}
        \centering
        \includegraphics[width=\textwidth]{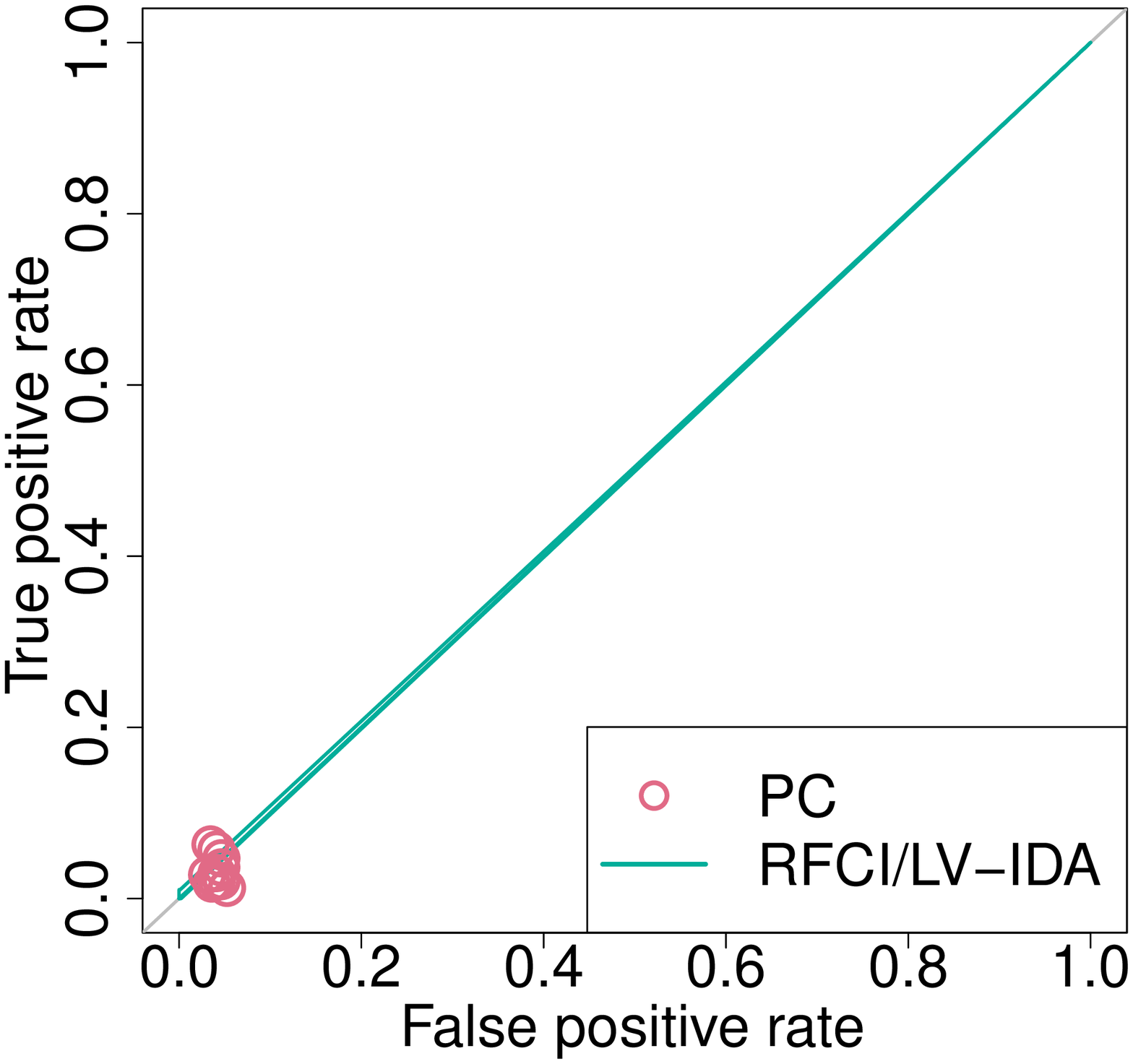}
        \caption{$\alpha = 0.1$}
        \label{fig:app-vary-alpha-3}
    \end{subfigure}
    \begin{subfigure}[b]{0.23\textwidth}
        \centering
        \includegraphics[width=\textwidth]{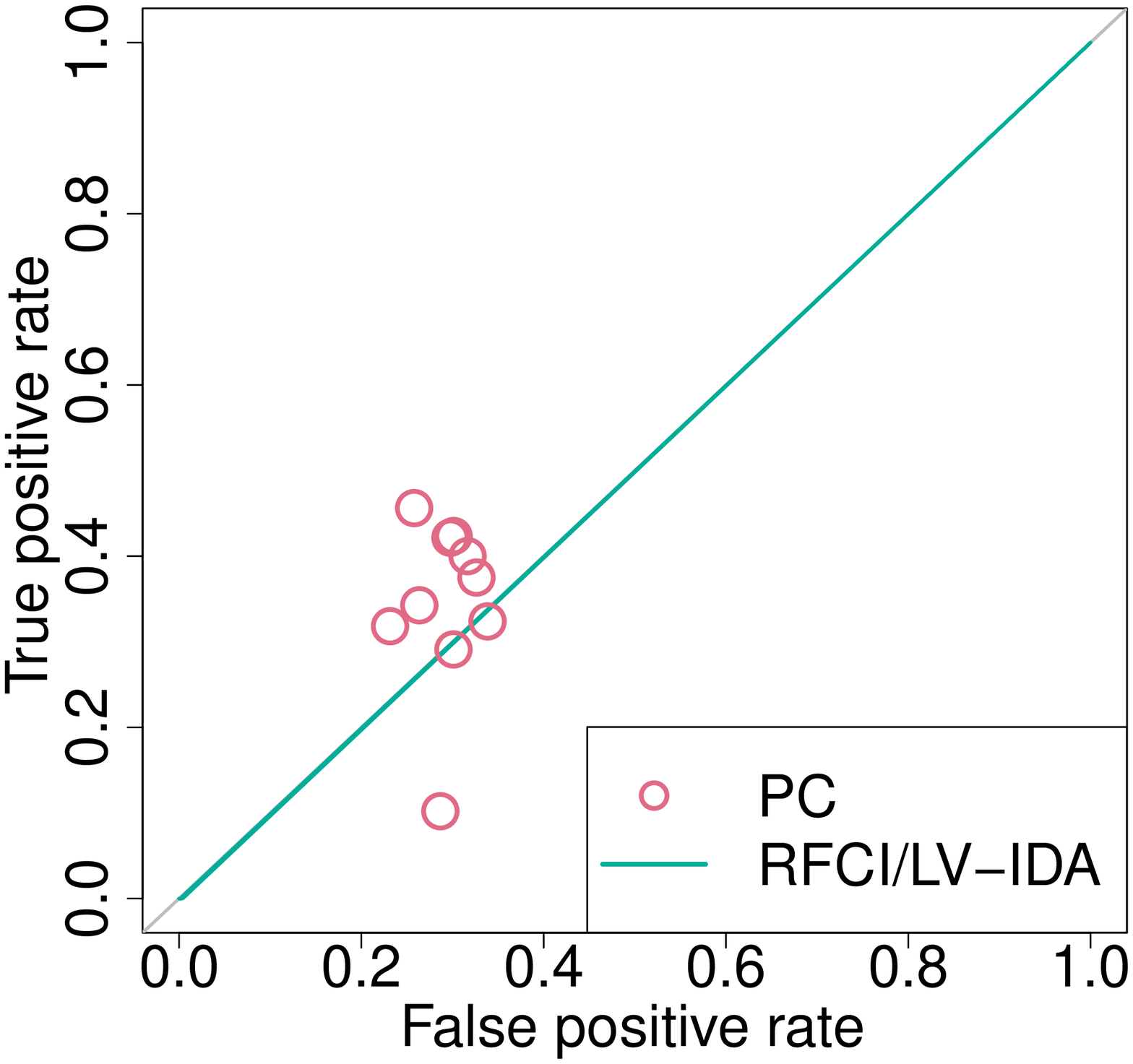}
        \caption{$\alpha = 0.5$}
        \label{fig:app-vary-alpha-4}
    \end{subfigure}
    
    \vspace{0.3cm}
    \hdashrule[0.5ex]{0.96\textwidth}{0.25pt}{2pt}
    \vspace{0.3cm}
    
    $p = 1000$ \\
    \begin{subfigure}[b]{0.23\textwidth}
        \centering
        \includegraphics[width=\textwidth]{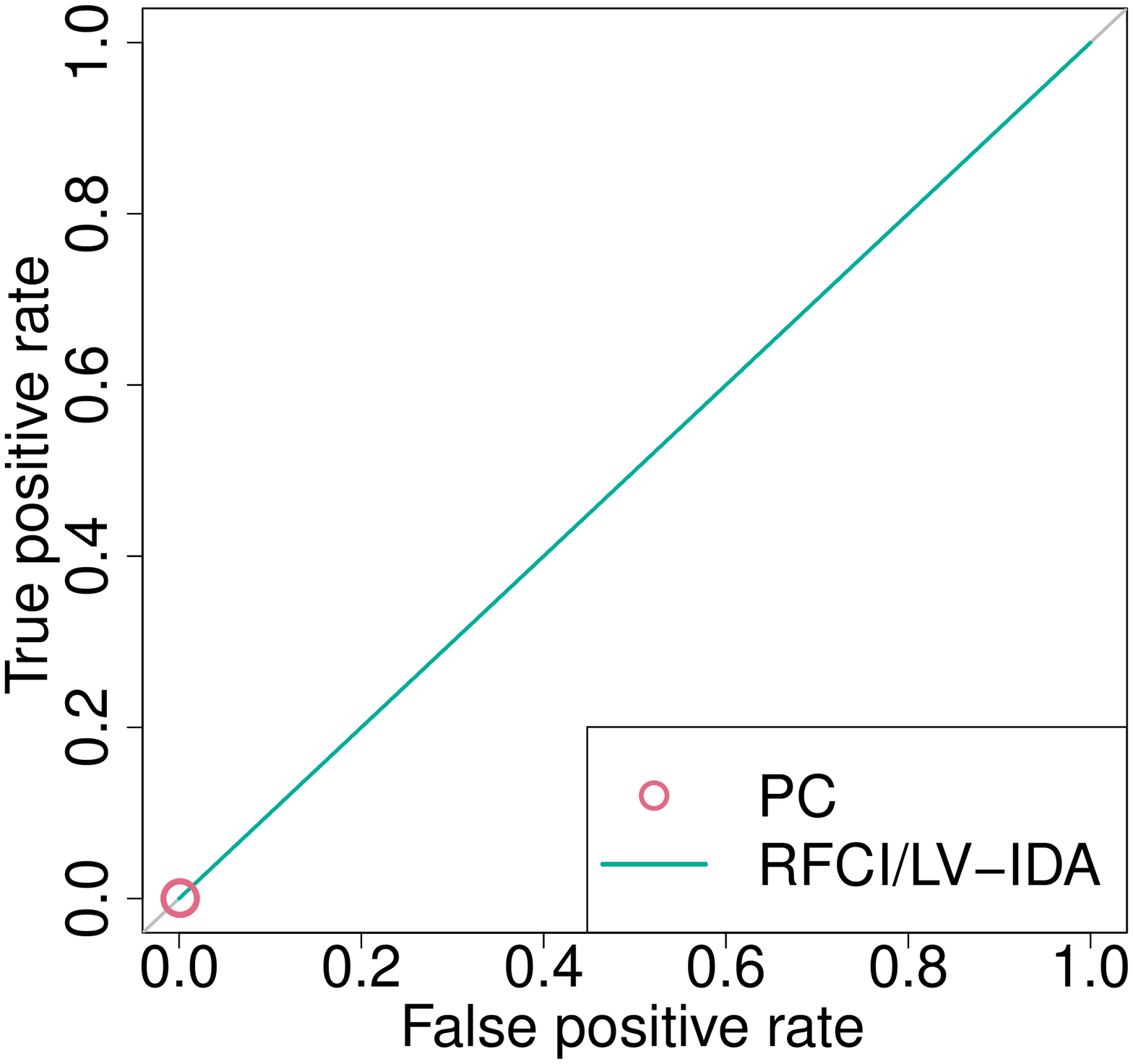}
        \caption{$\alpha = 0.01$}
        \label{fig:app-vary-alpha-5}
    \end{subfigure}
    \begin{subfigure}[b]{0.23\textwidth}
        \centering
        \vspace{-.4cm}
        \includegraphics[width=\textwidth]{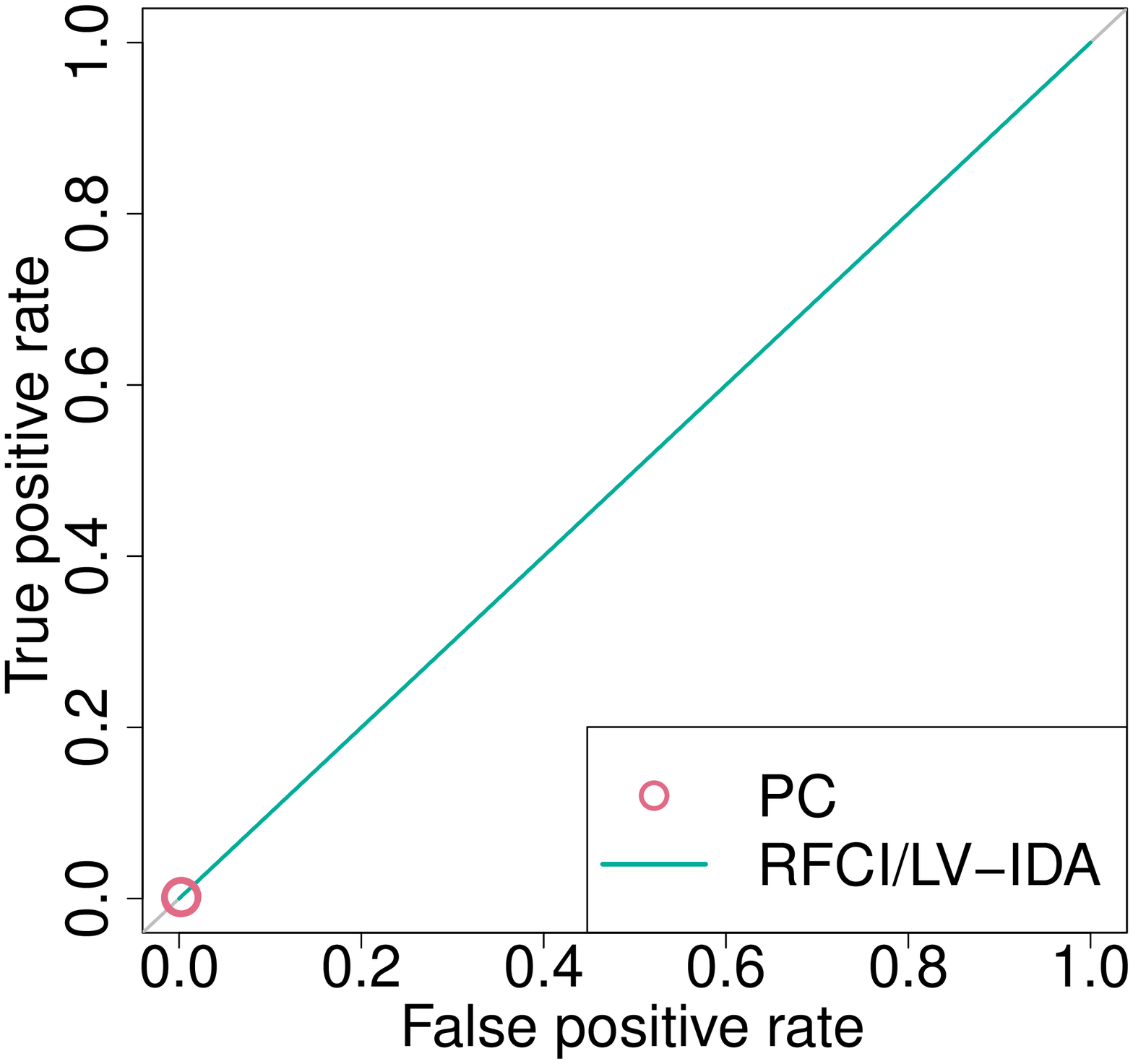}
        \caption{$\alpha = 0.05$}  
        \label{fig:app-vary-alpha-6}
    \end{subfigure}
    \begin{subfigure}[b]{0.23\textwidth}
        \centering
        \includegraphics[width=\textwidth]{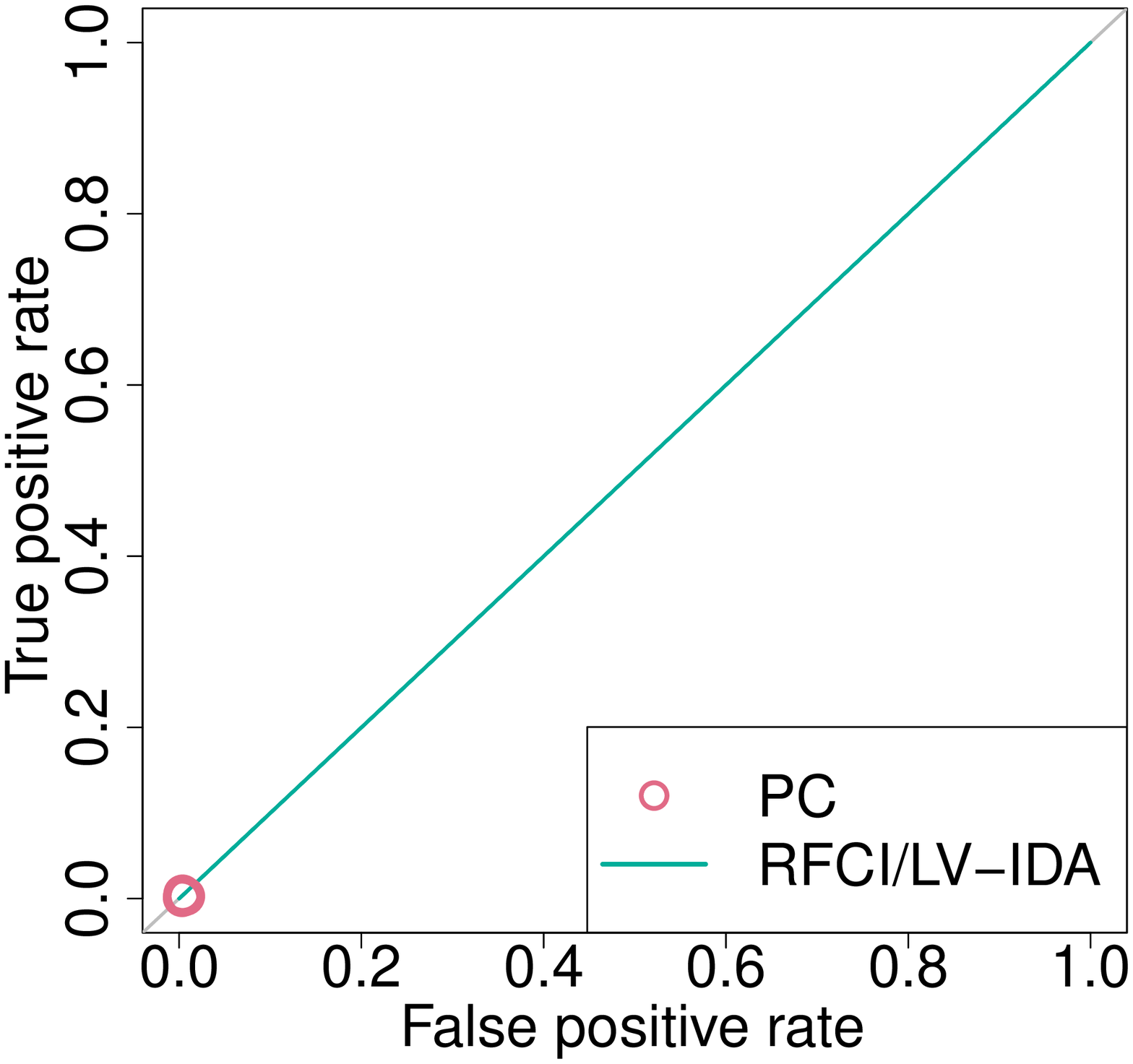}
        \caption{$\alpha = 0.1$}
        \label{fig:app-vary-alpha-7}
    \end{subfigure}
    \begin{subfigure}[b]{0.23\textwidth}
        \centering
        \includegraphics[width=\textwidth]{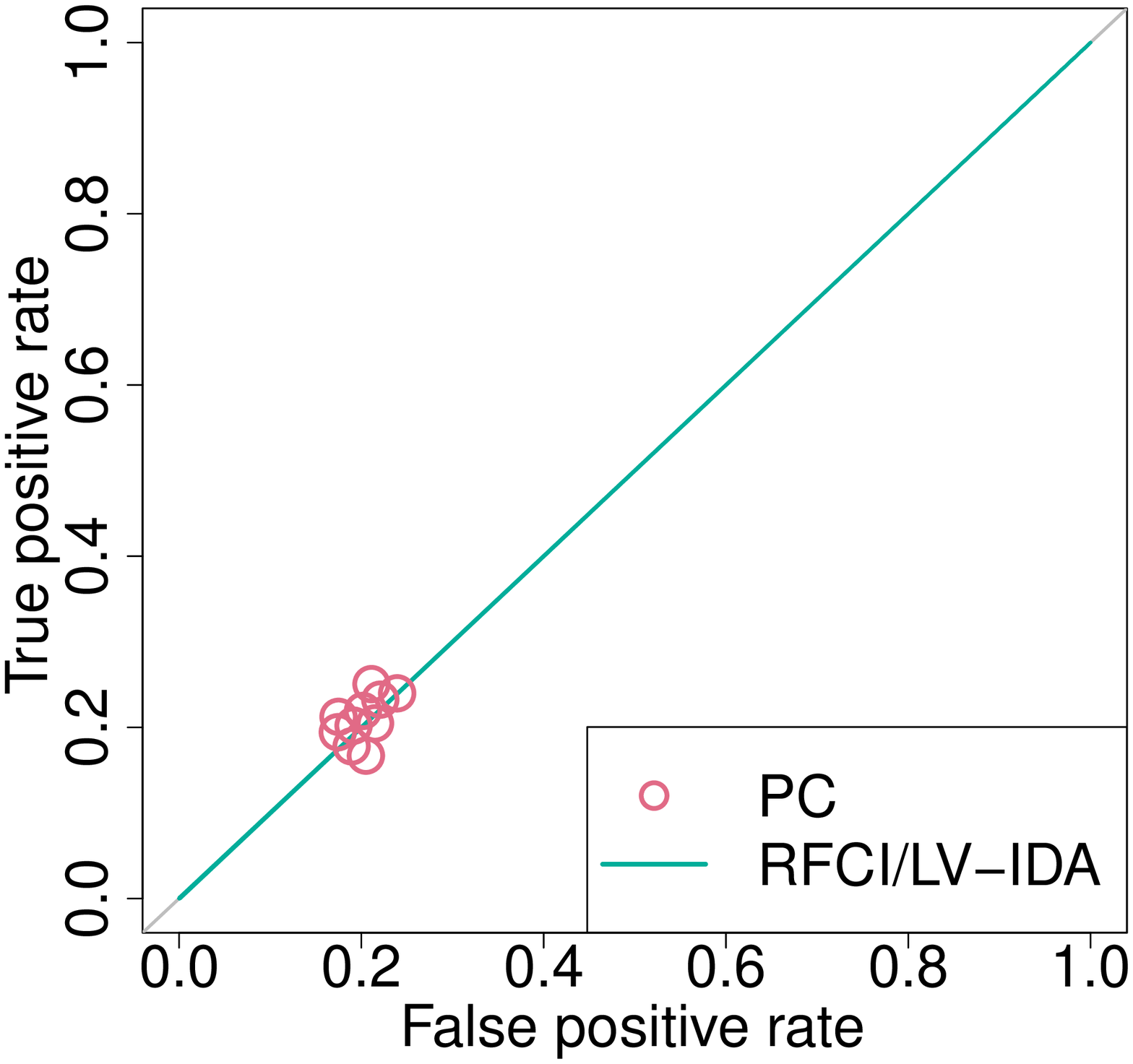}
        \caption{$\alpha = 0.5$}
        \label{fig:app-vary-alpha-8}
    \end{subfigure}
    
    \caption{The effect of varying  significance level $\alpha$ for PC and RFCI/LV-IDA. Results show the ROC curves and points on the ROC plane for 10 sampling iterations.}
    \label{fig:app-vary-alpha}
\end{figure}

\begin{figure}[tbh]
    \centering
    \includegraphics[width = 0.6\textwidth]{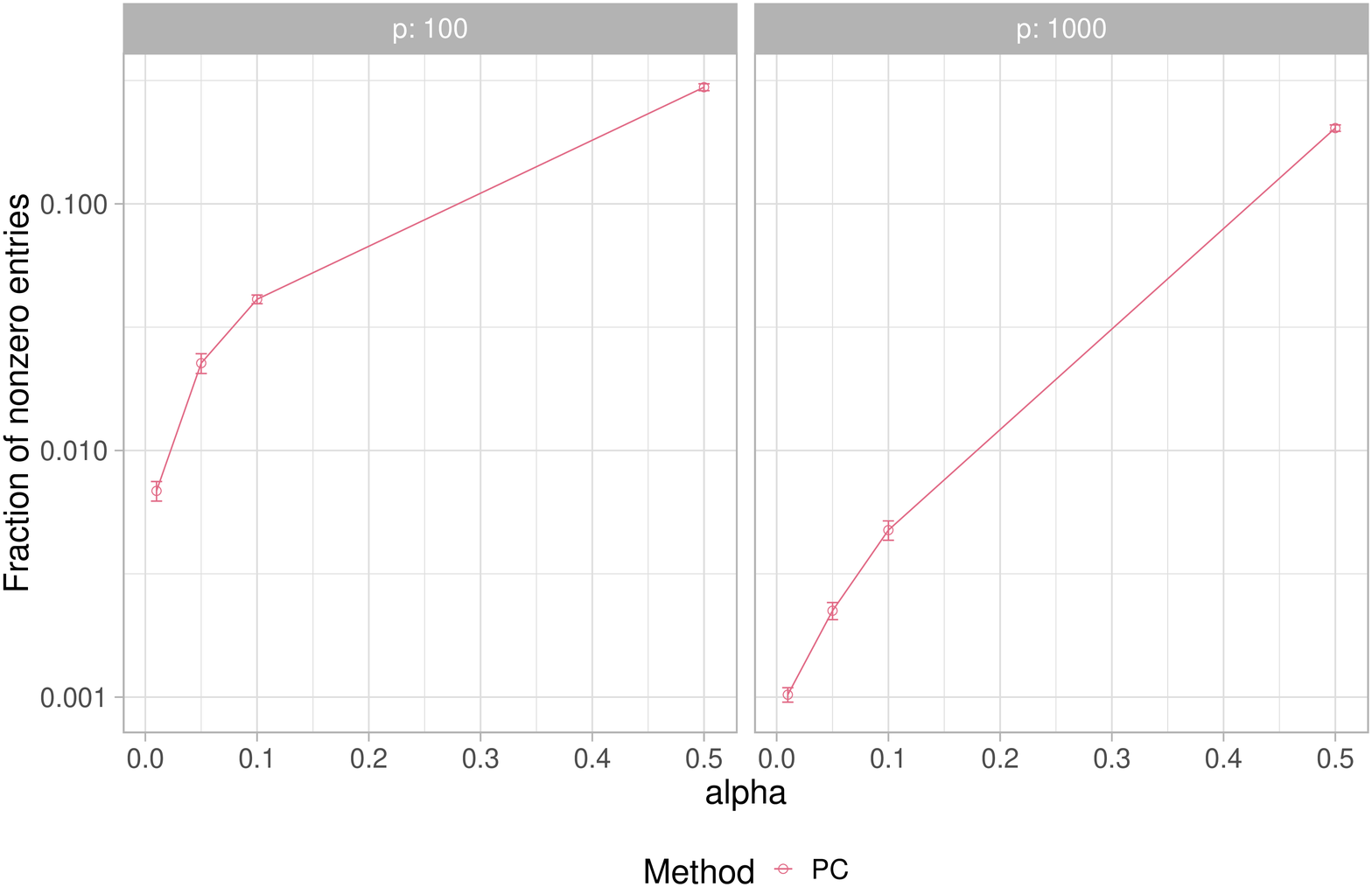}
    \caption{Fraction of nonzero entries in the estimated graph from the PC algorithm. Results show the mean over 10 sampling iterations with error bars indicating one standard error.}
    \label{fig:app-vary-alpha-sparsity-pc}
\end{figure}

\begin{figure}[tbh]
    \centering
    \includegraphics[width = 0.6\textwidth]{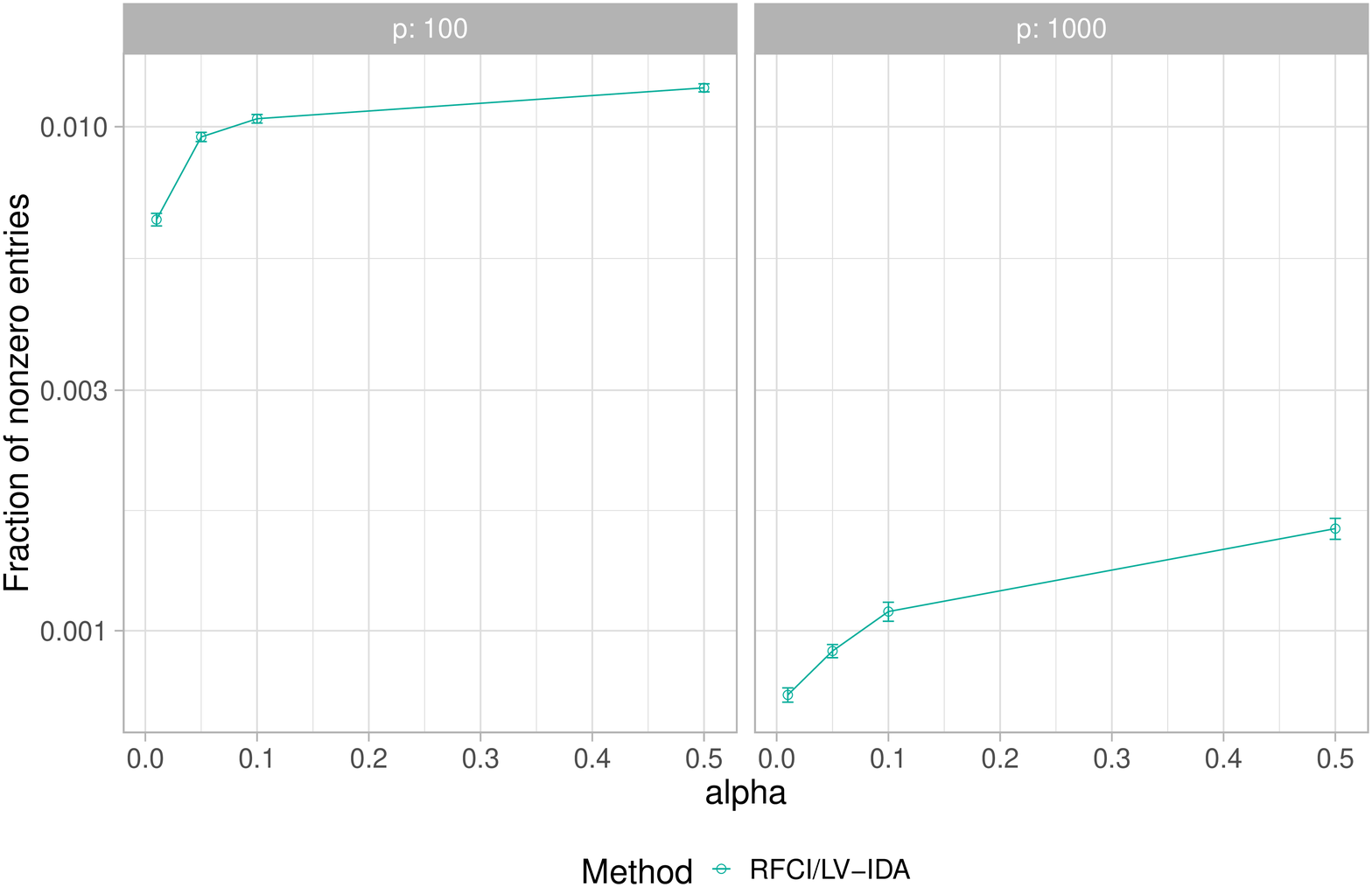}
    \caption{Fraction of nonzero entries in the estimated graph from the RFCI/LV-IDA algorithm. Results show the mean over 10 sampling iterations with error bars indicating one standard error.}
    \label{fig:app-vary-alpha-sparsity-lvida}
\end{figure}


\begin{figure}[tbh]
    \centering
    \includegraphics[width = 0.8\textwidth]{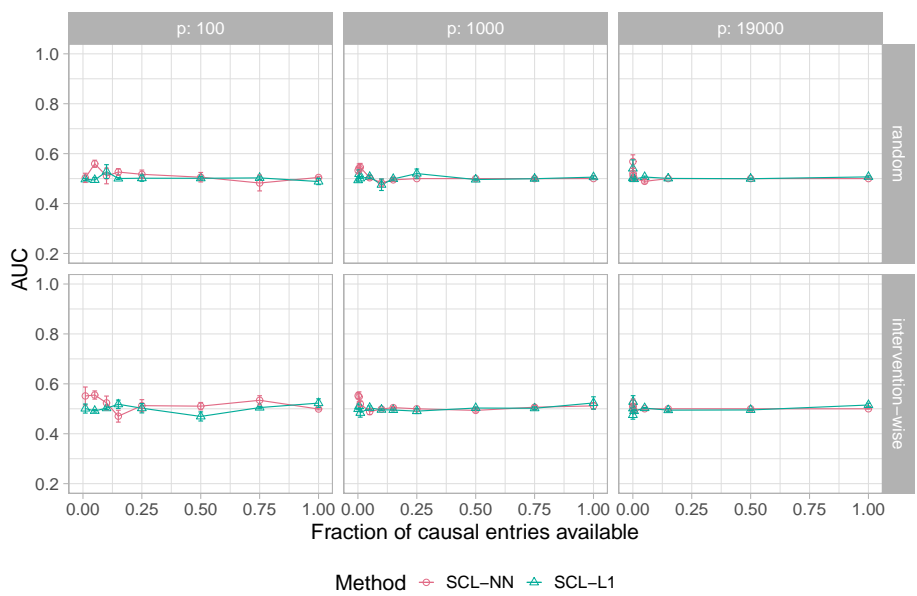}
    \caption{Learning using a small number of {\it entirely incorrect} ancestral relationships. This is a counterpart to Main Text Figure~\ref{fig:sparse-labelling}, with the ``positive" examples located entirely at random. Results show the mean over 10 sampling iterations with error bars indicating one standard error.}
    \label{fig:app-sparse-labelling-random}
\end{figure}